\documentclass[a4paper,11pt]{article}

\usepackage[english]{babel}
\usepackage[utf8]{inputenc}
\usepackage{amsmath}
\usepackage{graphicx}

\date{}
\title{Black-Box Generalization: Stability of Zeroth-Order Learning}

\author{Konstantinos E. Nikolakakis\textsuperscript{$\dagger$} \\\texttt{\small konstantinos.nikolakakis@yale.edu}
  \and  Farzin Haddadpour \\\texttt{\small farzin.haddadpour@yale.edu}
        \and \!\!\!Dionysios S. Kalogerias \\\texttt{\small dionysis.kalogerias@yale.edu} \and \qquad Amin Karbasi \\\qquad\texttt{\small amin.karbasi@yale.edu}     } 

\usepackage[T1]{fontenc}
\usepackage{lmodern}
\usepackage{cite}
\usepackage{appendix}

\PassOptionsToPackage{hyphens}{url}\usepackage{hyperref}%
\hypersetup{
    colorlinks=true,
    linkcolor=black,
    filecolor=magenta,      
    urlcolor=cyan,
    citecolor=blue	
}
\usepackage[hyphens]{url}
\usepackage{hyperref}
\usepackage[hyphenbreaks]{breakurl} 

\usepackage{booktabs}       
\usepackage{amsfonts}       
\usepackage{nicefrac}       
\usepackage{microtype}      
\usepackage{verbatim}
\usepackage{amsmath}
\usepackage{amssymb}
\usepackage{stackrel}
\usepackage{verbatim}
\usepackage{url}
\usepackage{algpseudocode}
\usepackage{graphicx}
\usepackage{enumerate}
\usepackage{algorithm}
\usepackage{algpseudocode}
\usepackage{float}
\usepackage{blindtext}
\usepackage{multirow}
\usepackage{dsfont}
\usepackage{mathtools}

\newcommand{\qedwhite}{\hfill \ensuremath{\Box}}

\newcommand\numberthis{\addtocounter{equation}{1}\tag{\theequation}}

\newcommand{\lp}{\left(}
\newcommand{\rp}{\right)}

\newcommand{\mc}{\mathcal}

\newcommand{\mbb}{\mathbb}

\usepackage[normalem]{ulem}
\newcommand{\cn}{\textcolor{red}{[\raisebox{-0.2ex}{\tiny\shortstack{citation\\[-1ex]needed}}]}}

\newcommand{\kn}[1]{\textcolor{blue}{\textbf{KN:} #1}}
\newcommand{\dk}[1]{\textcolor{red}{\textbf{DK:} #1}}
\newcommand{\fh}[1]{\textcolor{brown}{\textbf{FH:} #1}}
\newcommand\redout{\bgroup\markoverwith{\textcolor{red}{\rule[.5ex]{2pt}{0.4pt}}}\ULon}

\newcommand{\MDK}{\Gamma^d_K}
\renewcommand{\P}{\mbb{P}}
\newcommand{\E}{\mbb{E}}

\newcommand{\isnonconvex}{\mathds{1}_{f(\cdot)\notin\mc{C}}}
\newcommand{\Xmark}{{\Huge \text{\xmark} }}
\def\checkmark{\tikz\fill[scale=0.8](0,.35) -- (.25,0) -- (1,.7) -- (.25,.15) -- cycle;}

\newcommand{\eps}{\epsilon}

\newcommand{\Vast}{\bBigg@{5}}

\usepackage{enumitem}
\usepackage{amssymb}
\usepackage{pifont}
\newcommand{\cmark}{\text{\ding{51}}}
\newcommand{\xmark}{\text{\ding{55}}}
\usepackage{tikz}
\setitemize{noitemsep}
 \usepackage{makecell}

\usepackage{mathtools}
\DeclarePairedDelimiterX{\inp}[2]{\langle}{\rangle}{#1, #2}
\usepackage{geometry}
\allowdisplaybreaks
\newcolumntype{P}[1]{>{\centering\arraybackslash}p{#1}}
\newcolumntype{M}[1]{>{\centering\arraybackslash}m{#1}}
\newcommand{\BlackBox}{\rule{1.5ex}{1.5ex}}  

\newtheorem{theorem}{Theorem}
\newtheorem{lemma}[theorem]{Lemma}

\newtheorem{corollary}[theorem]{Corollary}
\newtheorem{definition}[theorem]{Definition}

\begin{document}
\maketitle

\vspace{-0.5cm}
\begin{abstract}%
We provide the first generalization error analysis for black-box learning through derivative-free optimization. Under the assumption of a Lipschitz and smooth unknown loss, we consider the \textit{Zeroth-order Stochastic Search} {(ZoSS)} algorithm, 
that updates a $d$-dimensional model by replacing stochastic gradient directions with stochastic differences of $K+1$ perturbed loss evaluations per dataset (example) query.
For both unbounded and bounded possibly nonconvex losses, we present the first generalization bounds for the ZoSS algorithm. These bounds coincide with those for SGD, and rather surprisingly are independent of $d$, $K$ and the batch size $m$, under appropriate choices of a slightly decreased learning rate. For bounded nonconvex losses and a batch size $m=1$, we additionally show that both generalization error and learning rate are independent of $d$ and $K$, and remain essentially the same as for the SGD, even for two function evaluations. 
Our results extensively extend and consistently recover established results for SGD in prior work,
on both generalization bounds and corresponding learning rates. If additionally $m=n$, where $n$ is the dataset size, we derive generalization guarantees for full-batch GD as well. 
\end{abstract}
\hspace{+0.5cm}\vspace{-0.1cm}\textbf{Keywords:}  Generalization Error, Zeroth-Order Optimization, Black-Box Learning

\section{Introduction}\footnotetext{\textsuperscript{$\dagger$}{Lead \& corresponding author}}\label{intro}
Learning 
methods often rely on empirical risk minimization objectives that highly depend on a limited training data-set. Known gradient-based approaches such as SGD train and generalize effectively in reasonable time~\cite{hardt2016train}. In contrast, emerging applications such as convex bandits~\cite{NIPS2011_67e103b0,shamir2017optimal,akhavan2020exploiting}, black-box learning~\cite{chen2017zoo}, federated learning~\cite{ZOFL}, reinforcement learning~\cite{vemula2019contrasting,kumar2021actor}, learning linear quadratic regulators~\cite{malik2019derivative,mohammadi2020linear}, and hyper-parameter tuning~\cite{nevergrad} 
 stand in need of \textit{gradient-free learning algorithms}~\cite{nevergrad,rios2013derivative,duchi2015optimal,nesterov2017random} due to an unknown loss/model or impossible gradient evaluation.

Given two or more function evaluations, zeroth-order algorithms~(see, e.g.,~\cite{nesterov2017random,balasubramanian2021zeroth}) aim to estimate the true gradient for evaluating and updating model parameters (say, of dimension $d$). In particular, Zeroth-order Stochastic Search (ZoSS)~\cite[Corollary 2]{duchi2015optimal},~\cite[Algorithm 1]{haghifam2021towards} uses $K+1$ function evaluations ($K\geq 1$), while deterministic zeroth-order approaches~\cite[Section 3.3]{chen2017zoo} require at least $K\geq d+1$ queries. 
The optimization error of the ZoSS algorithm is optimal as shown in prior work for convex problems~\cite{duchi2015optimal}, and suffers at most a factor of $\sqrt{d/K}$ in the convergence rate as compared with SGD. In addition to the optimization error, the importance of generalization error raises the question of how well zeroth-order algorithms generalize to unseen examples. In this paper, we show that the generalization error of ZoSS essentially coincides with that of SGD, under the choice of a slightly decreased learning rate. Assuming a Lipschitz and smooth loss function, we establish generalization guarantees for ZoSS, by extending stability-based analysis for SGD~\cite{hardt2016train}, to the gradient-free setting. In particular, we rely on the celebrated result that uniform algorithmic stability implies generalization~\cite{hardt2016train,bousquet2002stability,shalev2010learnability}.

Early works~\cite{bousquet2002stability,Devroye79distribution-freeperformance,kearns1999algorithmic,mukherjee2006learning,NIPS2000_49ad23d1} first introduced the notion of stability, and the connection between (uniform) stability and generalization. Recently, alternative notions of stability and generalization gain attention such as locally elastic stability~\cite{deng2021toward}, VC-dimension/flatness measures~\cite{jiang2019fantastic}, distributional stability~\cite{10.1145/2897518.2897566,NEURIPS2018_77ee3bc5,10.1145/3406325.3465358}, information theoretic bounds ~\cite{haghifam2021towards,steinke2020reasoning,pmlr-v54-alabdulmohsin17a,xu2017information,bu2020tightening,pmlr-v134-neu21a,harutyunyan2021information} mainly based on assuming a sub-Gaussian  loss, as well as connections between differential privacy and generalization~\cite{he2020robustness,pmlr-v161-he21a,yang2021stability,wang2021differentially}. 

In close relation to our paper, Hardt et al.~\cite{hardt2016train} first showed uniform stability final-iterate bounds for vanilla SGD. More recent works develop alternative generalization error bounds based on  high-probability analysis~\cite{NEURIPS2018_05a62416,feldman2019high,madden2020high,klochkov2021stability} and data-dependent variants~\cite{kuzborskij2018data}, or under different assumptions than those of prior works such as as strongly quasi-convex~\cite{gower2019sgd}, non-smooth convex~\cite{NIPS2016_8c01a759,bassily2020stability,pmlr-v119-lei20c,lei2021generalizationMP}, and pairwise losses~\cite{lei2021generalization,NEURIPS2020_f3173935}. In the nonconvex case,~\cite{zhou2021understanding} provide bounds that involve on-average variance of the stochastic gradients. Generalization performance of other algorithmic variants lately gain further attention, including SGD with early momentum~\cite{ramezani2021generalization}, randomized coordinate descent~\cite{wang2021stability}, look-ahead approaches~\cite{zhou2021towards}, noise injection methods~\cite{xing2021algorithmic}, and stochastic gradient Langevin dynamics ~\cite{pensia2018generalization,mou2018generalization,li2019generalization,NEURIPS2019_05ae14d7,zhang2021stability,farghly2021time,wang2021optimizing,wang2021analyzing}.

Recently, stability and generalization of full-bath GD has also been studied; see, e.g.,~\cite{NIPS2017_a5e0ff62, charles2018stability,amir2021never,amir2021sgd,richards2021stability}. 
In particular, Charles and Papailiopoulos.~\cite{charles2018stability} showed instability of GD for nonconvex losses. Still, such instability does not imply a lower bound on the generalization error of GD (in expectation).
In fact, Hoffer et al.~\cite{NIPS2017_a5e0ff62} showed empirically that the generalization of GD is not affected by the batch-size, and for large enough number of iterations GD generalizes comparably to SGD. Our analysis agree with the empirical results of Hoffer et al.~\cite{NIPS2017_a5e0ff62}, as we show that (for smooth losses) the generalization of ZoSS (and thus of SGD) is independent of the batch size.
\paragraph{Notation.} \vspace{-0.4cm} 
We denote the training data-set $S$ of size $n$ as $\{z_i\}^n_{i=1}$, where $z_i$ are i.i.d. observations of a random variable $Z$ with unknown distribution $\mc{D}$. The parameters of the model are vectors of dimension $d$, denoted by $W\in \mbb{R}^d$, and $W_t$ is the output at time $t$ of a (randomized) algorithm $A_S$. The (combined) loss function $f(\cdot,z):\mbb{R}^d \rightarrow \mbb{R}^+$ is uniformly Lipschitz and smooth for all $z\in\mc{Z}$. We denote the Lipschitz constant as $L$ and the smoothness parameter by $\beta$. The number of function (i.e., loss) evaluations (required at each iteration of the ZoSS algorithm) is represented by $K+1\in \mbb{N}$. We denote by $\Delta f$ the smoothed approximation of the loss gradient, associated with parameter $\mu$. The parameter $\MDK \triangleq \sqrt{(3d-1)/K}+1$ prominently appears in our results. We denote the gradient of the loss function with respect to model parameters $W$, by $\nabla f(W ,z)\equiv\nabla_{w}f(w ,z)|_{w=W}$. We denote the mini batch at $t$ by $J_t$, and $m\triangleq |J_t|$.

\begin{table}[t]\label{table:results}
\centering
\begin{tabular}{|>{\centering}m{3.8cm}||>{\centering}m{6.78cm}|c|c|c|}
\hline 
\multicolumn{5}{|c|}{Generalization Error Bounds: ZoSS vs SGD}\tabularnewline
\hline 
\hline 
Algorithm & Bound & NC & UB & MB\tabularnewline
\hline 
ZoSS (\textbf{this work})\\
$\alpha_{t}\le C/(t\Gamma_{K}^{d})$ & \vspace{-5bp}
$\dfrac{1+(CB)^{-1}}{n}\big((2+c)CL^{2}\big)^{\frac{1}{C\beta+1}}(eT)^{\frac{C\beta}{C\beta+1}}$ & $\cmark$ &$\xmark$ & $\xmark$\tabularnewline
\hline 
SGD, $\alpha_{t}\le C/t$\\
Hardt et al.~\cite{hardt2016train} & \vspace{-4bp}
$\dfrac{1+(CB)^{-1}}{n}\big(2CL^{2}\big)^{\frac{1}{C\beta+1}}(eT)^{\frac{C\beta}{C\beta+1}}$ & $\cmark$ & $\xmark$ & $\xmark$\tabularnewline
\hline 
ZoSS (\textbf{this work})\\
$\alpha_{t}\le C/t$ & \vspace{2bp}
$\dfrac{3e\big(1+(CB)^{-1}\big)^{2}}{2n}\big(1+(2+c)CL^{2}\big)T$\\
\vspace{2bp}
(\textit{independent of both $d$ and $K$})\vspace{2bp}
 & $\cmark$ & $\xmark$ & $\xmark$ \tabularnewline
\hline 
\hline 
ZoSS (\textbf{this work})

$\alpha_{t}\le\frac{\log\Big(1+\frac{C\beta}{\Gamma_{K}^{d}}\sqrt{\frac{3d-1}{K}}\Big)}{T\beta\sqrt{(3d-1)/K}}$ & $\dfrac{(2+c)CL^{2}}{n}$ & $\xmark$ & $\cmark$ &$\cmark$\tabularnewline
\hline 
SGD, $\alpha_{t}\le C/T$\\
Hardt et al.~\cite{hardt2016train} & \vspace{-3.5bp}
$\dfrac{2CL^{2}}{n}$ & $\xmark$ & $\cmark$ &$\cmark$\tabularnewline
\hline 
\hline 
ZoSS (\textbf{this work}) \\
$\alpha_{t}\le C/(T\Gamma_{K}^{d})$ & \vspace{-1bp}
$\dfrac{(2+c)L^{2}(e^{C\beta}-1)}{n\beta}$ & $\cmark$ & $\cmark$ &$\cmark$\tabularnewline
\hline 
\vspace{-9bp}
ZoSS (\textbf{this work}) \\
$\alpha_{t}\le\frac{\log(1+C\beta)}{T\beta\Gamma_{K}^{d}}$ & \vspace{2bp}
$\dfrac{(2+c)CL^{2}}{n}$\\
\vspace{2bp}
(proper choice of $C$ in previous bound)\vspace{2bp}
 & $\cmark$ & $\cmark$ &$\cmark$\tabularnewline
\hline 
ZoSS (\textbf{this work}) \\
$\alpha_{t}\le C/(t\Gamma_{K}^{d})$\vspace{1bp}
 & \vspace{2bp}
$\dfrac{(2+c)L^{2}(eT)^{C\beta}}{n}\min\{C+\beta^{-1},C\log(T)\}$\vspace{2bp}
 & $\cmark$ & $\cmark$ & $\cmark$\tabularnewline
\hline 
\end{tabular}

\caption{ 
A list of the generalization error bounds developed herein for ZoSS (Eq. \ref{eq:ZO_updates}) in comparison with SGD, with 
$\mu\leq   cL\MDK/n \beta (3+d)^{3/2} $, for $c>0$. 
In the table, \textquotedblleft NC\textquotedblright{} and \textquotedblleft UB\textquotedblright{}
stand for \textquotedblleft nonconvex\textquotedblright{} and \textquotedblleft unbounded\textquotedblright ,
respectively. \textquotedblleft MB\textquotedblright  ~corresponds to the mini-batch algorithm and for any batch size. Also, $\alpha_t$ denotes the stepsize of ZoSS/SGD, and $T$ the total number of iterations.\label{table_nonconvex}}

\end{table}
\subsection{Contributions}
Under the assumption of Lipschitz and smooth loss functions, we provide generalization guarantees for black-box learning, extending the analysis of prior work by Hardt et al.~\cite{hardt2016train} to the gradient free setting.
In particular, we establish uniform stability and generalization error bounds for the final iterate of the ZoSS algorithm; 
see Table \ref{table:results} for a summary of the results. In more detail, the contributions of this work are as follows: 
\begin{itemize}[leftmargin=10pt]
\item For unbounded \textit{and} bounded losses, we show generalization error bounds identical to SGD, with a slightly decreased learning rate. 
 Specifically, the generalization error bounds are independent of the dimension $d$, the number of evaluations $K$ and the batch-size $m$. Further, a large enough number of  evaluations ($K$) provide fast generalization even in the high dimensional regime.
\item For bounded nonconvex losses and single (example) query updates ($m=1$), we show that both the ZoSS generalization error and  learning rate are independent of $d$ and $K$, similar to that of SGD~\cite[Theorem 3.8]{hardt2016train}. This property guarantees efficient generalization even with two function evaluations.
\item In the full information regime (i.e., when the number of function evaluations $K$ grow to $\infty$), the ZoSS generalization bounds also provide guarantees for SGD by recovering the results in prior work~\cite{hardt2016train}. Further, we derive novel SGD bounds for unbounded nonconvex losses, as well as mini-batch SGD for any batch size. Our results subsume generalization guarantees for full-batch ZoSS and GD algorithms.   
\end{itemize}

\section{Problem Statement}\label{Problem Statement}
Given a data $S\triangleq \{z_i\}^n_{i=1}$ of i.i.d samples $z_i$ from an unknown distribution $\mc{D}$, our goal is to find the parameters $w^*$ of a learning model such that $ w^* \in \arg \min_{w} R(w)$, where $R(w)\triangleq \E_{Z\sim \mc{D}} [ f(w,Z) ]$.
%
%
Since the distribution $\mc{D}$ is not known, we consider the empirical risk 
\begin{align}
    R_S (w) \triangleq \frac{1}{n} \sum^n_{i=1} f(w , z_i),
\end{align} 
and the corresponding empirical risk minimization (ERM) problem to find $w^*_s \in \arg \min_{w} R_S (w)$. For a (randomized) algorithm $A_S$ with input $S$ and output $W=A(S)$, the excess risk $\eps_{\text{excess}}$ is bounded by the sum of the generalization error $\eps_{\mathrm{gen}}$ and the optimization error $\eps_{\mathrm{opt}}$,  \begin{align}
    \eps_{\mathrm{excess}} \triangleq \E_{S,A}[R(W)]  - R(w^*) = \underbrace{\E_{S,A}[R(W) -R_S (W) ]}_{\eps_{\mathrm{gen}}} + (\underbrace{\E_{S,A}[R_S (W)] - R(w^*)}_{\eps_{\mathrm{opt}}}).
\end{align} 
To analyze and control $\eps_{\mathrm{gen}}$, we prove uniform stability bounds which imply generalization~\cite[Theorem 2.2]{hardt2016train}. Specifically, if for all i.i.d. sequences $S,S'\in \mc{Z}^n$ that differ in one entry, we have $\sup_z \E_{A} [f(A(S),z) - f(A(S'),z)]\leq \eps_{\text{stab}}$,
for some $\eps_{\text{stab}}>0 $, then $\eps_{\mathrm{gen}}\leq \eps_{\text{stab}}$. Because the loss is $L$-Lipschitz, $\eps_{\text{stab}}$ may then be chosen as $L\sup_{S,S'}\mathbb{E}_A\Vert A(S) -A(S') \Vert$.

Our primary goal in this work is to develop uniform stability bounds for a \textit{gradient-free} algorithm $A_S$ of the form $w_{t+1}= w_t - \alpha_t \Delta f_{w_t,z}$, where $\Delta f_{w_t,z}$ only depends on loss function evaluations. To achieve this without introducing unnecessary assumptions, we consider a novel algorithmic stability error decomposition approach. In fact, the stability error introduced at time $t$ by $A_S$ breaks down into the stability error of SGD and an approximation error due to missing gradient information. Let $G_t(\cdot)$ and $G'_t(\cdot)$ be the following SGD update rules \begin{align}
    G_t(w)&\triangleq w-\alpha_t  \nabla f(w ,z_{i_t}), \quad G'_t(w)\triangleq  w-\alpha_t  \nabla f(w ,z'_{i_t}),\label{eq:GradientGt}
\end{align} under inputs $S,S'$ respectively, and let $i_t \in\{1,2,\ldots,n\}$ be a random index chosen uniformly and independently by the random selection rule of the algorithm, for all $t\leq T$. 
Similarly we use the notation $\tilde{G}(\cdot)$ and $\tilde{G}'(\cdot)$ to denote the iteration mappings of $A_S$, i.e., 
\begin{align}
\tilde{G}_t(w)&\triangleq w - \alpha_t {\Delta f}_{w,z_{i_t}},\quad \tilde{G}'_t(w)\triangleq w - \alpha_t {\Delta f}_{w,z'_{i_t}}.\label{eq:ZOGt}
\end{align}
Then, as we also discuss later on (Lemma \ref{lemma:ZoSS Growth Recursion}), the iterate stability error $  \tilde{G}_t(w) - \tilde{G}'_t(w')$ of $A_S$, for any $w,w'\in\mbb{R}^d$ and for all at $t\leq T$, may be decomposed as \begin{align*}
& \tilde{G}_t(w) - \tilde{G}'_t(w') \propto  \underbrace{ G_{t}(w) -  G'_{t}(w') }_{\eps_{\text{GBstab}}} + \underbrace{\big[ \nabla f(w ,z_{i_t}) - {\Delta f}_{w,z_{i_t}} \big] + \big[\nabla f(w' ,z'_{i_t}) - {\Delta f}_{w',z'_{i_t}} \big]}_{\eps_{\text{est}}},\numberthis\label{eq:error_decomposition}
\end{align*} where $\eps_{\text{GBstab}}$ denotes the gradient-based stability error (associated with SGD), and $\eps_{\text{est}}$ denotes the gradient approximation error. We now proceed by formally introducing ZoSS.

\section{Zeroth-Order Stochastic Search (ZoSS)}
As a \textit{gradient-free} alternative of the classical SGD algorithm, we consider the ZoSS scheme, with  iterates generated according to the following (single-example update) rule %
\begin{align}
    W_{t+1} &= W_t - \alpha_t \frac{1}{K}\sum^K_{k=1}\frac{f(W_t +\mu U^t_{k},z_{i_t})-f(W_t ,z_{i_t} )}{\mu}U^t_{k}, \quad U^t_{k} \sim \mc{N}(0 ,I_d), \quad \mu\in \mbb{R}^+,\label{eq:ZO_updates}
\end{align}where $\alpha_t\ge0$ is the corresponding learning rate (for the mini-batch update rule we refer the reader to Section \ref{Section mini_batch}). At every iteration $t$, ZoSS generates $K$ i.i.d. standard normal random vectors $U^t_{k}, k=1,\ldots,K$, 
and obtains $K+1$ loss evaluations on perturbed model inputs. Then ZoSS evaluates a smoothed approximation of the gradient for some $\mu>0$. In light of the discussion in Section \ref{Problem Statement}, we define the ZoSS smoothed gradient step at time $t$ as
\begin{align} \label{eq:ZoSS_directional}
  {\Delta f}^{K,\mu}_{w,z_{i_t}} \equiv 
  {\Delta f}^{K,\mu,\mathbf{U}^t}_{w,z_{i_t}}&\triangleq \frac{1}{K}\sum^K_{k=1} \frac{ f(w +\mu U^t_{k},z_{i_t})-f(w ,z_{i_t} )}{\mu} U^t_{k}.
\end{align}

\subsection{ZoSS Stability Error Decomposition} To show stability bounds for ZoSS, we decompose its error into two parts through the stability error decomposition discussed in Section \ref{Problem Statement}. Under the ZoSS update rule, Eq.~\eqref{eq:error_decomposition} holds by considering the directions ${\Delta f}_{w,z_{i_t}}$ and ${\Delta f}_{ w',z'_{i_t}} $ according to ZoSS smoothed approximations \eqref{eq:ZoSS_directional}. 
Then for any $w,w'\in\mbb{R}^d$, the iterate stability error $ \tilde{G}_{t}(w) -  \tilde{G}'_{t}(w')$ of ZoSS at $t$, breaks down into the gradient based error $\eps_{\text{GBstab}}$ and approximation error $\eps_{\text{est}}$.

The error term $\eps_{\text{GBstab}}$ expresses the stability error of the gradient based mappings~\cite[Lemma 2.4]{hardt2016train} and inherits properties related to the SGD update rule. The error $\eps_{\text{est}}$ captures the approximation error of the ZoSS smoothed approximation and depends on $K$ and $\mu$. The consistency of the smoothed approximation with respect to SGD follows from $\lim_{K \uparrow \infty, \mu \downarrow 0} {\Delta f}^{K,\mu}_{w,z}  = \nabla f (w , z) $ for all $w\in\mbb{R}$ and $z\in\mc{Z}$. Further, the stability error is also consistent since $\lim_{K \uparrow \infty, \mu \downarrow 0} |\eps_{\text{est}}| = 0 $. Later on, we use the ZoSS error decomposition in Eq. \eqref{eq:error_decomposition} together with a variance reduction lemma (Lemma \ref{lemma:variance_reduction_zero_mean}), to derive exact expressions on the iterate stability error $ \tilde{G}_{t}(w) -  \tilde{G}'_{t}(w')$ for fixed $K$ and $\mu>0$ (see Lemma \ref{lemma:ZoSS Growth Recursion}). 
Although in this paper we derive stability bounds and bounds on the $\eps_{\mathrm{gen}}$, the excess risk $\eps_{\text{excess}}$ depends on both errors $\eps_{\text{gen}}$ and $\eps_{\text{opt}}$. In the following section, we briefly discuss known results on the $\eps_{\text{opt}}$ of zeroth-order methods, including convex and nonconvex losses. 
\subsection{Optimization Error in Zeroth-Order Stochastic Approximation}
Convergence rates of the ZoSS optimization error and related zeroth-order variants have been extensively studied in prior works; see e.g.,~\cite{nesterov2017random,balasubramanian2021zeroth,nesterov2011random}. For the convex loss setting, when $K+1$ function evaluations are available and no other information regarding the loss is given, the ZoSS algorithm achieves optimal rates with respect to the optimization error $\eps_{\text{opt}}$. Specifically, under the assumption of a closed and convex loss, Duchi et al.~\cite{duchi2015optimal} provided a lower bound for the minimax convergence rate and showed that $\eps_{\text{opt}}=\Omega (\sqrt{d/K}) $, for any algorithm that approximates the gradient given $K+1$ evaluations. In the nonconvex setting Ghadimi et al.~\cite{ghadimi2013stochastic,ghadimi2016mini} established sample complexity guarantees for the zeroth-order approach to reach an approximate stationary point.

\section{Main Results}\label{Sec_Results}
For our analysis, we introduce the same assumptions on the loss function (Lipschitz and smooth) as appears in prior work~\cite{hardt2016train}. Additionally, we exploit the $\eta$-expansive and $\sigma$-bounded properties of the SGD mappings $G_t(\cdot)$ and $G'_t(\cdot)$ in Eq. \eqref{eq:GradientGt}.\footnote{\cite[Definition 2.3]{hardt2016train}: An update rule $G(\cdot)$ is $\eta$-expansive if $\Vert G(w) -G(w') \Vert \leq \eta \Vert w-w' \Vert$ for all
$w,w'\in\mbb{R}^d$. If $\Vert w-G(w)\Vert\leq \sigma $ then it is $\sigma$-bounded.} The mappings $G_t(\cdot)$ and $G'_t(\cdot)$ are introduced for analysis purposes due to the stability error decomposition given in Eq. \eqref{eq:error_decomposition} and no further assumptions or properties are required for the zeroth-order update rules $\tilde{G}_t (\cdot)$ and $\tilde{G}'_t (\cdot)$ given in Eq. \eqref{eq:ZOGt}. Recall that the $\eta$-expansivity of $G_t(\cdot)$ holds for $\eta=1+\beta\alpha_t$ if the loss is nonconvex, and $\eta=1$ if the loss is convex and $\alpha_t\leq 2/\beta$~\cite[Lemma 3.6]{hardt2016train}. 
Note that $G_t(\cdot)$ is always $\sigma$-bounded for $\sigma= L\alpha_t$~\cite[Lemma 3.3.]{hardt2016train}. 
\subsection{Stability Analysis}\label{SubSec_Lemmas}
We start by providing a lemma that is useful in the proofs of our main results. 
    \begin{lemma}\label{lemma:variance_reduction_zero_mean}
Let $\mathbf{U}_k \in \mbb{R}^d, k\in \{1,2\ldots,K\}$ be i.i.d standard Gaussian.\footnote{Similar bounds (i.e., $\mc{O}(\sqrt{d/K})$) hold for other distributions as well, e.g., when the $\mathbf{U}_k$'s are uniformly distributed (and independent) in $[-1,+1]$.}
For every random vector $\mathbf{V} \in \mbb{R}^d$ independent of all $\mathbf{U}_k, k\in \{1,2\ldots, K\}$, it is true that
\begin{align}
     &\E \left[ \Bigg\Vert \frac{1}{K} \sum^K_{k=1} \inp{\mathbf{V}}{ \mathbf{U}_k}  \mathbf{U}_k - \mathbf{V} \Bigg\Vert \Bigg| \mathbf{V} \right] 
     \leq \sqrt{\frac{3d-1}{K}} \Vert\mathbf{V} \Vert.
\end{align} 
\end{lemma} Different versions of Lemma \ref{lemma:variance_reduction_zero_mean} appear already in prior works (see, e.g.,~\cite[Proof of Corollary 2]{duchi2015optimal}. For completeness and clarity we provide a short proof of Lemma \ref{lemma:variance_reduction_zero_mean} in the Appendix, Section \ref{App_Proof_Variance_reduction}. 
Exploiting Lemma \ref{lemma:variance_reduction_zero_mean}, we show a growth recursion lemma for the iterates of the ZoSS.
\begin{lemma}[ZoSS Growth Recursion]\label{lemma:ZoSS Growth Recursion}
Consider the sequences of updates $\{\tilde{G}_{t}\}^T_{t=1}$ and $\{\tilde{G}'_{_t}\}^T_{t=1}$. Let $w_0 =w'_0$ be the starting point, $w_{t+1} = \tilde{G}_{t} (w_t) $ and $w'_{t+1} = \tilde{G}'_{t} (w'_t) $ for any $t\in \{1,\ldots , T\}$. Then for any $w_t,w'_t\in\mbb{R}^d$ and $t\geq 0$ the following recursion holds
\begin{align*}
   \E [\Vert \tilde{G}_{t} (w_t) - \tilde{G}'_{t} (w'_t) \Vert] \leq  
     \begin{cases}
      \Big(\eta+ \alpha_t  \sqrt{\frac{3d-1}{K}}  \beta \Big) \Vert w_t - w'_t \Vert   + \mu \beta\alpha_t  (3+d)^{3/2}, &\text{ if }\tilde{G}_{t}(\cdot) =\tilde{G}'_{t}(\cdot), \\
       \Vert w_t - w'_t \Vert + 2\alpha_t L \MDK    +   \mu \beta\alpha_t  (3+d)^{3/2}, &\text{ if } \tilde{G}_{t}(\cdot) \neq \tilde{G}'_{t}(\cdot).
     \end{cases}
\end{align*}
\end{lemma} The growth recursion of ZoSS characterizes the stability error that it is introduced by the ZoSS update and according to the outcome of the randomized selection rule at each iteration. Lemma \ref{lemma:ZoSS Growth Recursion} extends growth recursion results for SGD in prior work~\cite[Lemma 2.5]{hardt2016train} to the setting of the ZoSS algorithm. If $K\rightarrow \infty$ and $\mu\rightarrow 0$ (while the rest of the parameters are fixed), then $\MDK\rightarrow 1$, and the statement recovers that of the SGD~\cite[Lemma 2.5]{hardt2016train}. 
\paragraph{Proof of Lemma \ref{lemma:ZoSS Growth Recursion}.}
Let $S$ and $S'$ be two samples of size $n$ differing in only a single example, and let $\tilde{G}_t (\cdot),\tilde{G}'_t(\cdot)$ be the update rules of the ZoSS for each of the sequences $S,S'$ respectively. First under the event $\mathcal{E}_t\triangleq \{ \tilde{G}_t(\cdot) \equiv \tilde{G}'_t(\cdot)\}$ (see Eq. \eqref{eq:ZOGt}), 
by 
applying the Taylor expansion there exist vectors $W^*_{k,t}$ and $W^\dagger_{k,t}$ with $j^{\text{th}}$ coordinates in the intervals $\big( w^{(j)}_t , w^{(j)}_t+\mu U^{(j)}_{k,t}\big) \cup\big(  w^{(j)}_t+\mu U^{(j)}_{k,t}, w^{(j)}_t \big)$ and $\big( w'^{(j)}_t , w'^{(j)}_t+\mu U^{(j)}_{k,t}\big)\cup\big(  w'^{(j)}_t+\mu U^{(j)}_{k,t}, w'^{(j)}_t \big)$, respectively, such that for any $w_t ,w'_t \in\mbb{R}^d$ we have
\begin{align*}
    & \!\!\!\!\tilde{G}_t(w_t) - \tilde{G}'_t(w'_t) = \tilde{G}_t(w_t) - \tilde{G}_t(w'_t) \\
    &\!\!\!\! =  w_t - w'_t  -   \frac{\alpha_t}{K}\sum^K_{k=1} \inp{\nabla f(w_t ,z_{i_t})-\nabla f(w'_t ,z_{i_t})}{ U^t_k}U^t_k \numberthis\label{eq:taylor1}  \\&\!\!\!\!\quad - \frac{\alpha_t}{K}\sum^K_{k=1} \lp \frac{\mu}{2} U^{\text{T}}_k \nabla^2_{w}f(w ,z_{i_t})|_{w=W^*_{k,t}} U^t_k \rp U^t_k   + \frac{\alpha_t}{K}\sum^K_{k=1} \lp \frac{\mu}{2} U^{\text{T}}_k \nabla^2_{w}f(w ,z_{i_t})|_{w=W^\dagger_{k,t}} U^t_k\rp U^t_k  \\
    &\!\!\!\! = \underbrace{w_t-\alpha_t  \nabla f(w_t ,z_{i_t})}_{G(w_t)} -  \underbrace{\lp w'_t  -\alpha_t\nabla f(w'_t ,z_{i_t})\rp}_{G'(w'_t)\equiv G(w'_t)}  \\&\!\!\!\!\quad - \frac{\alpha_t}{K}\sum^K_{k=1} \lp \frac{\mu}{2} U^{\text{T}}_k \nabla^2_{w}f(w ,z_{i_t})|_{w=W^*_{k,t}} U^t_k \rp U^t_k   + \frac{\alpha_t}{K}\sum^K_{k=1} \lp \frac{\mu}{2} U^{\text{T}}_k \nabla^2_{w}f(w ,z_{i_t})|_{w=W^\dagger_{k,t}} U^t_k\rp U^t_k  \\ & \quad  -   \alpha_t\bigg( \frac{1}{K}\sum^K_{k=1} \inp{\nabla f(w_t ,z_{i_t})-\nabla f(w'_t ,z_{i_t})}{ U^t_k}U^t_k- ( \nabla f(w_t ,z_{i_t})-\nabla f(w'_t ,z_{i_t}))\bigg)\numberthis\label{eq:base_diff_1}. 
\end{align*} We find \eqref{eq:base_diff_1} by adding and subtracting $\alpha_t  \nabla f(w_t ,z_{i_t})$ and $\alpha_t  \nabla f(w'_t ,z_{i_t})$ in Eq. \eqref{eq:taylor1}. 
Recall that $U^t_k$ are independent for all $k\leq K$, $t\leq T$ and that the mappings $G(\cdot)$ and $G'(\cdot)$ defined in Eq. \eqref{eq:base_diff_1}, are $\eta$-expansive. The last display and the triangle inequality give \begin{align*}
&\!\!\!\! \E [\Vert \tilde{G}_{t} (w_t) - \tilde{G}_{t} (w'_t) \Vert]\\&\!\!\!\!\!\!\leq  \Vert G(w_t) - G(w'_t) \Vert \! +\! \frac{2\alpha_t }{K}\sum^K_{k=1} \frac{\mu \beta }{2} \E \left[ \Vert U^t_k\Vert ^3 \right]
\!+ \!\alpha_t  \sqrt{\frac{3d-1}{K}} \E[\Vert \nabla f(w_t ,z_{i_t})-\nabla f(w'_t ,z_{i_t}) \Vert] \numberthis \label{eq:Lemma1ongradientdifferences}\\
&\!\!\!\!\!\leq  \eta  \Vert w_t - w'_t \Vert   +  \frac{2\alpha_t }{K}\sum^K_{k=1} \frac{\mu \beta }{2} \E \left[ \Vert U^t_k\Vert ^3 \right] + \alpha_t  \sqrt{\frac{3d-1}{K}}  \beta  \Vert w_t - w'_t \Vert \numberthis \label{eq:expansive_smooth}\\
 &\!\!\!\!\!\leq\lp \eta+ \alpha_t  \sqrt{\frac{3d-1}{K}}  \beta \rp  \Vert w_t - w'_t \Vert   + \mu \beta\alpha_t  (3+d)^{3/2}, \numberthis \label{eq:iddU}
\end{align*} where \eqref{eq:Lemma1ongradientdifferences} follows from \eqref{eq:base_diff_1} and Lemma \ref{lemma:variance_reduction_zero_mean}, 
and for \eqref{eq:expansive_smooth} we applied the $\eta$-expansive property of $G(\cdot)$ (see~\cite[Lemma 2.4 and Lemma 3.6]{hardt2016train}) and the $\beta$-smoothness of the loss function.\footnote{For all $z\in\mc{Z}$ and $W\in\mbb{R}^d$ it is true that $\Vert \nabla^2_{w}f(w ,z)|_{w=W} \Vert \leq \beta $.} Finally \eqref{eq:iddU} holds since the random vectors $U^t_k\sim\mc{N}(0,I_d)$ are identically distributed for all $k\in \{1,2,\ldots,K\}$ and $\E \Vert U^t_k\Vert ^3\leq (3+d)^{3/2}$. Eq. \eqref{eq:iddU} gives the first part of the recursion.

Similar to \eqref{eq:base_diff_1}, under the event $\mathcal{E}^c_t\triangleq \{ \tilde{G}_t(\cdot) \neq \tilde{G}'_t(\cdot)\}$, we find \begin{align*}
    & \tilde{G}_t(w_t) - \tilde{G}'_t(w'_t) 
    \\ & = \underbrace{w_t-\alpha_t  \nabla f(w_t ,z_{i_t})}_{G(w_t)} -  \underbrace{\lp w'_t  -\alpha_t\nabla f(w'_t ,z'_{i_t})\rp}_{G'(w'_t)}  \\&\quad - \frac{\alpha_t}{K}\sum^K_{k=1} \lp \frac{\mu}{2} U^{\text{T}}_k \nabla^2_{w}f(w ,z_{i_t})|_{w=\tilde{W}^*_{k,t}} U^t_k \rp U^t_k   + \frac{\alpha_t}{K}\sum^K_{k=1} \lp \frac{\mu}{2} U^{\text{T}}_k \nabla^2_{w}f(w ,z'_{i_t})|_{w=\tilde{W}^\dagger_{k,t}} U^t_k\rp U^t_k  \\ & \quad  -   \alpha_t \bigg( \frac{1}{K}\sum^K_{k=1} \inp{\nabla f(w_t ,z_{i_t})-\nabla f(w'_t ,z'_{i_t})}{ U^t_k}U^t_k - ( \nabla f(w_t ,z_{i_t})-\nabla f(w'_t ,z'_{i_t}))\bigg). \numberthis\label{eq:W_difference_Ec}
\end{align*} By using the last display, triangle inequality, Lemma \ref{lemma:variance_reduction_zero_mean} and $\beta$-smoothness, we find\begin{align*}
& \E [\Vert \tilde{G}_{t} (w_t) - \tilde{G}_{t} (w'_t) \Vert]\\
&\leq  \Vert G(w_t) - G'(w'_t) \Vert   +  \frac{2\alpha_t }{K}\sum^K_{k=1} \frac{\mu \beta }{2} \E  [\Vert U^t_k\Vert ^3]  + \alpha_t  \sqrt{\frac{3d-1}{K}} \E [\Vert \nabla f(w_t ,z_{i_t})\!-\!\nabla f(w'_t ,z'_{i_t}) \Vert ] \\
&\leq  \min\{\eta,1\} \delta_t + 2\sigma_t    +  \frac{2\alpha_t }{K}\sum^K_{k=1} \frac{\mu \beta }{2}\E [ \Vert U^t_k\Vert ^3 ] + 2 L \alpha_t  \sqrt{\frac{3d-1}{K}}  \numberthis \label{eq:expansive_smoothsecond}\\
 &\leq \delta_t + 2\alpha_t L \MDK    +   \mu \beta\alpha_t  (3+d)^{3/2} , \numberthis \label{eq:iddUsecond}
\end{align*} where \eqref{eq:expansive_smoothsecond} follows from the triangle inequality and $L-$Lipschitz condition, while the upper bound on $ \Vert G(w_t) - G'(w'_t) \Vert$ comes from~\cite[Lemma 2.4]{hardt2016train}. Finally, \eqref{eq:iddUsecond} holds since $\eta\geq 1$ for both convex and nonconvex losses, $\sigma_t=L\alpha_t$ and $\E \Vert U^t_k\Vert ^3\leq (3+d)^{3/2}$ for all $k\in\{1,\ldots,K\}$. This shows the second part of recursion. \qedwhite
\begin{definition}\label{def:delta_t_0}
Let $\mc{I}$ be an adapted stopping time that corresponds to the first iteration index that the single distinct instance of the two data-sets $S,S'$ is sampled by ZoSS. For any $t_0\in \{0,1,\ldots,n\}$ we define the event  
$
    \mc{E}_{\delta_{t_0}} \triangleq \{\mc{I}>t_0 \} \equiv \{ \delta_{t_0} =0  \}.
$
\end{definition}
Recall that $W_T$ and $W_{T'}$ are the outputs of ZoSS \eqref{eq:ZO_updates} with input data sequences $S$ and $S'$, respectively. The next result provides a stability bound on the expected value of the norm $\Vert  W_{T} - W'_{T} \Vert$.
\begin{lemma}[\textbf{ZoSS Stability | Nonconvex Loss}]\label{thm:recur_solution_nonconvex}
Assume that the loss function $f(\cdot, z)$ is $L$-Lipschitz and $\beta$-smooth for all $z\in \mc{Z}$. Consider the ZoSS algorithm \eqref{eq:ZO_updates} with final-iterate estimates $W_T$ and $W'_T$, corresponding to the data-sets $S,S'$, respectively (that differ in exactly one entry). Then the discrepancy $\delta_T \triangleq \Vert  W_{T} - W'_{T} \Vert$, under the event $\mc{E}_{\delta_{t_0}}$, satisfies the inequality
\begin{align}
   \!\! &\E[\delta_T|\mc{E}_{\delta_{t_0}}]\leq \lp \frac{2 L}{n} \MDK    +   \mu \beta (3+d)^{3/2} \rp  \sum^T_{t=t_0 +1} \alpha_t \prod^T_{j=t+1} \lp 1 + \beta \alpha_j \MDK\lp 1-\frac{1}{n}\rp\rp.\label{eq:delta_bounds}
\end{align}
\end{lemma} The corresponding bound of Lemma \ref{thm:recur_solution_nonconvex} for convex losses is slightly tighter than the bound in \eqref{eq:delta_bounds}. Since the two bounds differ only by a constant, the consequent results of Lemma \ref{thm:recur_solution_nonconvex} are essentially identical for convex losses as well. We provide the equivalent version of Lemma \ref{thm:recur_solution_nonconvex} for convex losses in Appendix \ref{Appendix_Complementary_results}. 

\paragraph{Proof of Lemma \ref{thm:recur_solution_nonconvex}.} Consider the events $\mathcal{E}_t\triangleq \{ \tilde{G}_t(\cdot) \equiv \tilde{G}'_t(\cdot)\}$ and $\mathcal{E}^c_t\triangleq \{ \tilde{G}_t(\cdot) \neq \tilde{G}'_t(\cdot)\}$ (see Eq. \eqref{eq:ZOGt}). Recall that $\P(\mathcal{E}_t)=1-1/n$ and $\P(\mathcal{E}^c_t)=1/n$ for all $t\leq T$. For any $t_0\geq 0$, a direct application of Lemma \ref{lemma:ZoSS Growth Recursion} gives
\begin{align*}
    \E[\delta_{t+1}|\mc{E}_{\delta_{t_0}}] &= \P (\mc{E}_t)\E[\delta_{t+1}|\mc{E}_t,\mc{E}_{\delta_{t_0}}] + \P (\mc{E}^c_t)\E[\delta_{t+1}|\mc{E}^c_t,\mc{E}_{\delta_{t_0}}] \\
    &=\lp 1-\frac{1}{n}\rp\E[\delta_{t+1}|\mc{E}_t,\mc{E}_{\delta_{t_0}}] + \frac{1}{n}\E[\delta_{t+1}|\mc{E}^c_t ,\mc{E}_{\delta_{t_0}}]\\
    & \leq   \lp \eta+ \alpha_t \beta \sqrt{\frac{3d-1}{K}}+ \frac{1}{n}\lp 1- \eta- \alpha_t \beta \sqrt{\frac{3d-1}{K}} \rp \rp \E[ \delta_t|\mc{E}_{\delta_{t_0}}]   \\ & \quad  + \frac{2\alpha_t L}{n} \MDK    +   \mu \beta\alpha_t  (3+d)^{3/2}.\numberthis\label{eq:recusrion} 
\end{align*} 
Define \begin{align}
  R_t = R(\eta,\alpha_t,\beta , n, d,K) \triangleq \lp \eta+ \alpha_t \beta \sqrt{\frac{3d-1}{K}}+ \frac{1}{n}\lp 1- \eta- \alpha_t \beta \sqrt{\frac{3d-1}{K}} \rp \rp,\label{eq:R(.)}
\end{align} then by solving the recursion in \eqref{eq:recusrion} we find \begin{align}
    \E[\delta_T|\mc{E}_{\delta_{t_0}}] \leq \lp \frac{2 L}{n} \MDK    +   \mu \beta (3+d)^{3/2} \rp  \sum^T_{t=t_0+1} \alpha_t \prod^T_{j=t+1}R_j.\label{eq:Recur_sol_R}
\end{align} We consider the last inequality for nonconvex loss functions with $\eta=1+\beta\alpha_t$ and convex loss functions with $\eta=1$ to derive Lemma \ref{thm:recur_solution_nonconvex} and Lemma \ref{thm:recur_solution_convex} respectively (Appendix \ref{Appendix_Complementary_results}). \qedwhite
\subsection{Generalization Error Bounds}
For the first generalization error bound, we evaluate the right part of the inequality \eqref{eq:delta_bounds} for decreasing step size and bounded nonconvex loss. Then the Lipschitz condition provides a uniform stability condition for the loss and yields the next theorem.   
\begin{theorem}[\textbf{Nonconvex Bounded Loss | Decreasing Stepsize}]\label{thm:nonconvex-bounded-loss_tight}
Let $f(\cdot, z)\in [0,1]$ be $L$-Lipschitz and $\beta$-smooth for all $z\in \mc{Z}$. Consider the ZoSS update rule \eqref{eq:ZO_updates} with $T$ the total number of iterates, $\alpha_t \leq C/t \MDK $ for some (fixed) $C>0$ and for all $t\leq T$, and fixed $\mu\leq   cL\MDK/n \beta (3+d)^{3/2} $ for some $c>0$. Then the generalization error of ZoSS is bounded by
\begin{align*}
      \!\!\!\eps_{\mathrm{gen}}&\leq \frac{\lp  (2+c) CL^2     \rp^{\frac{1}{C\beta+1}}(eT)^\frac{C\beta}{C\beta+1}}{n}  \max\!\left\{\! 1, 1+(C\beta)^{-1}  \! -\! \frac{e^{\beta C}}{\beta C^{{\frac{1}{C\beta+1}}}} \!\lp \frac{(2+c)L^2}{e T}\rp^\frac{C\beta}{C\beta+1} \!  \right\}\numberthis\label{eq:tighter_main_result} \\
      &\leq \frac{\lp 1+(C\beta)^{-1}\rp\lp  (2+c) CL^2     \rp^{\frac{1}{C\beta+1}}}{n} (eT)^\frac{C\beta}{C\beta+1}.\label{eq:main_result_short}\numberthis 
\end{align*}
\end{theorem} Inequality \eqref{eq:tighter_main_result}, as a tighter version of \eqref{eq:main_result_short}, provides a meaningful bound in marginal cases, i.e., \begin{align}
       \lim_{\beta\downarrow 0 } \E \left[ |f(W_T,z) - f(W'_T,z)| \right] \leq \frac{(2+c)CL^2}{n} \max\left\{ \log \lp \frac{eT}{(2+c)CL^2} \rp , 1 \right\}.
   \end{align} By neglecting the negative term in \eqref{eq:tighter_main_result} we find \eqref{eq:main_result_short}, that is the ZoSS equivalent of SGD~\cite[Theorem 3.8]{hardt2016train}. When $K\rightarrow \infty$ and $c\rightarrow 0$, then $\MDK\rightarrow 1$, and the inequalities \eqref{eq:tighter_main_result}, \eqref{eq:main_result_short} reduce to a generalization bound for SGD. Inequality \eqref{eq:tighter_main_result} matches that of~\cite[Theorem 3.8]{hardt2016train}, and \eqref{eq:tighter_main_result} provides a tighter generalization bound for SGD as well.
%
%

Next, we provide a bound on the generalization error for nonconvex losses that comes directly from Theorem \ref{thm:nonconvex-bounded-loss_tight}. In contrast to Theorem \ref{thm:nonconvex-bounded-loss_tight}, the next result provides learning rate and a generalization error bounds, both of which are independent of the dimension and the number of function evaluations.
\begin{corollary}\label{thm:coroallry_dimensional_free}
Assume that the loss function $f(\cdot, z)\in [0,1]$ is $L$-Lipschitz and $\beta$-smooth for all $z\in \mc{Z}$. Consider the ZoSS update rule \eqref{eq:ZO_updates} with $\mu\leq cL\MDK/(n\beta (3+d)^{3/2}) $, $T$ the total number of iterates, and $\alpha_t \leq C/t $ for some (fixed) $C>0$ and for all $t\leq T$. Then the generalization error of ZoSS is bounded by \begin{align}
    \eps_{\mathrm{gen}} \leq \lp 1+(\beta C )^{-1}\rp^2 \lp 1 + (2+c)CL^2\rp \frac{3Te}{2n}.
\end{align}
\end{corollary} As a consequence, even in the high dimensional regime $d\rightarrow\infty$, two function evaluations (i.e., $K=1$) are sufficient for the ZoSS to achieve $\eps_{\mathrm{gen}} = \mc{O}(T/n)$, with the learning rate being no smaller than that of SGD. We continue by providing the proof of Theorem \ref{thm:nonconvex-bounded-loss_tight}. For the proof of Corollary \ref{thm:coroallry_dimensional_free}, see Appendix \ref{Appendix_proof_corollary}.
\paragraph{Proof of Theorem \ref{thm:nonconvex-bounded-loss_tight}.}We start by observing that $1-\eta - \alpha_t \beta \sqrt{\frac{3d-1}{K}} <0$ since $\eta\geq 1$. Eq. \eqref{eq:R(.)} and \eqref{eq:Recur_sol_R} give $R(\eta,\alpha_t,\beta , n, d,K) \leq ( \eta+ \alpha_t \beta \sqrt{(3d-1)/K} ) \triangleq \tilde{R}_t$, and 
\begin{align}
    \E[\delta_T|\mc{E}_{\delta_{t_0}}] \leq \lp \frac{2 L}{n} \MDK    +   \mu \beta (3+d)^{3/2} \rp  \sum^T_{t=t_0+1} \alpha_t \prod^T_{j=t+1}\tilde{R}_j .\label{eq:Recur_sol_R_simple}
\end{align} Recall that $\eta = 1 +\beta \alpha_t $ for general (nonconvex) losses (see~\cite{hardt2016train}). Assuming that $\alpha_t\leq C/t\MDK$ for all $t\leq T$, we have \begin{align*}
    \!\!\!\!\!\!\!\E[\delta_T|\mc{E}_{\delta_{t_0}}] &\leq \lp \frac{2 L}{n} \MDK    +   \mu \beta (3+d)^{3/2} \rp  \sum^T_{t=t_0+1} \alpha_t \prod^T_{j=t+1}\lp 1+ \alpha_j \beta\MDK \rp\\
    & \leq \frac{C}{\MDK}\lp \frac{2 L}{n} \MDK    +   \mu \beta (3+d)^{3/2} \rp  \sum^T_{t=t_0+1} \frac{1}{t} \prod^T_{j=t+1}\lp 1+ \frac{C\beta}{j} \rp \\
    &\leq \frac{C}{\MDK}\lp \frac{2 L}{n} \MDK    +   \mu \beta (3+d)^{3/2} \rp  \sum^T_{t=t_0+1} \frac{1}{t} \prod^T_{j=t+1} \exp\lp \frac{C\beta}{j} \rp \numberthis\label{eq:exponential} \\
    &\leq \frac{C (eT)^{\beta C}}{\MDK}\lp \frac{2 L}{n} \MDK    +   \mu \beta (3+d)^{3/2} \rp  \sum^T_{t=t_0+1} \frac{1}{t} \frac{1}{(t+1)^{\beta C}}\numberthis \label{eq:multiple_steps1}\\
    &\leq \underbrace{\beta^{-1}\lp\MDK\rp^{-1}\lp \frac{2 L}{n} \MDK    +   \mu \beta (3+d)^{3/2} \rp }_{D}\!\!\lp \lp \frac{eT}{t_0}\rp^{\beta C} \!\!\!- e^{\beta C}\rp.\numberthis \label{eq:output_sensitivity_nonconvex_dlr}
\end{align*} In the above, the inequality $1+x \leq e^x$ gives \eqref{eq:exponential}, inequality \eqref{eq:multiple_steps1} follows from the inequality $\sum^T_{j=t+1} 1/j \leq \log T - \log (t+1) + 1 $, and inequality \eqref{eq:output_sensitivity_nonconvex_dlr} comes from the next inequality and integral evaluation $\sum^T_{t=t_0+1}t^{-\beta C-1}\leq \int^T_{t=t_0} x^{-\beta C-1} dx= (\beta C)^{-1} (t_0^{-\beta C} - T^{-\beta C} ) $. We define $q \triangleq \beta C$ and find the value of $t_0$ that minimizes the right part of~\cite[Lemma 3.11]{hardtarxiv}\footnote{\cite[Lemma 3.11]{hardtarxiv} applies to the ZoSS update rule \eqref{eq:ZO_updates} similar to SGD for nonnegative and $L$-Lipschitz losses. Note that $\P (\mc{I}\leq t_0)\leq t_0/n$, $t_0\in \{0,\ldots, T\}$, and \eqref{eq:t_0_inequality} comes from the Lipschitz assumption on the loss as $\E \left[ |f(W_T,z) - f(W'_T,z)| \right]\leq  \P (\mc{I}\leq t_0) \E \big[ |f(W_T,z) - f(W'_T,z)| \big| \mc{E}^c_{\delta_{t_0}} \big] + L \E [\delta_{T}|\mc{E}_{\delta_{t_0}}]$.} \begin{align}
   \E \left[ |f(W_T,z) - f(W'_T,z)| \right] \leq  \frac{t_0}{n} \sup_{w,z}f(w,z) +L\E[\delta_T|\mc{E}_{\delta_{t_0}}] \leq  \frac{t_0}{n} + L D  \lp \lp \frac{eT}{t_0}\rp^q \!\!\!- e^q \rp,\label{eq:t_0_inequality}
\end{align} which is $t^*_0  =  \min\{ \lp q n LD \rp^{1/(q+1)} (eT)^{q/(q+1)} , T \}$. Then \eqref{eq:t_0_inequality} gives
\begin{align*}
   \!\!\!&\E \left[ |f(W_T,z) - f(W'_T,z)| \right] 
    \\&\leq\max\left\{ \frac{\lp q n LD \rp^{\frac{1}{q+1}} (eT)^\frac{q}{q+1}}{n}, \frac{1+1/q}{n} \lp q n LD \rp^{\frac{1}{q+1}} (eT)^\frac{q}{q+1}-LDe^q\right\}.\numberthis\label{eq:nonconvexbound_incl_mu}
   \end{align*} Choosing $\mu \leq c  L \MDK/n \beta (3+d)^{3/2} $ for some $c>0$ 
   in \eqref{eq:nonconvexbound_incl_mu} 
proves our claim.  \qedwhite

In the unbounded loss case, we apply Lemma \ref{thm:recur_solution_nonconvex} by setting $t_0 = 0$ (recall that $t_0$ is a free parameter, while the algorithm depends on the random variable $\mc{I}$). The next result provides a generalization error bound for the ZoSS algorithm with constant step size. In the first case of the theorem, we also consider the convex loss as a representative result, as we show the same bound holds for an appropriate choice of greater learning rate than the learning rate of the nonconvex case. The convex case for the rest of the results of this work can be similarly derived.
\begin{theorem}[\textbf{Unbounded Loss | Constant Step Size}]\label{thm:Nonconvex Convex Unbounded Loss case2 - Constant Step Size}
Assume that the loss $f(\cdot, z)$ is $L$-Lipschitz, $\beta$-smooth for all $z\in \mc{Z}$. Consider the ZoSS update rule \eqref{eq:ZO_updates} with $\mu\leq cL\MDK/(n\beta (3+d)^{3/2}) $ for some $c>0$. Let $T$ be the total number of iterates and for any $t\leq T$,
\vspace{-4pt}
\begin{itemize}[leftmargin=10pt]
    \item if $f(\cdot,z)$ is convex for all $z\in \mc{Z}$ and $\alpha_t \leq \min\{ \log\big(1+C\beta(1-1/\MDK)\big)/ T\beta (\MDK-1),2/\beta\} $, or if $f(\cdot,z)$ is nonconvex and $\alpha_t \leq\log\big(1+C\beta\big)/ T\beta \MDK $, for some $C>0$ then
    \begin{align}
    \eps_{\mathrm{gen}} \leq \dfrac{(2+c)C  L^2}{n},
\end{align}
\item if $f(\cdot,z)$ is nonconvex and $\alpha_t \leq C/ T \MDK$, for some $C>0$ then
\begin{align}
    \eps_{\mathrm{gen}} \leq  \frac{ L^2\lp 2     +   c \rp(e^{C\beta}-1) }{n\beta }.
\end{align}
\end{itemize}
\end{theorem} For the proof of Theorem \ref{thm:Nonconvex Convex Unbounded Loss case2 - Constant Step Size} see Appendix \ref{Appendix_proof_thm:Nonconvex Convex Unbounded Loss case2 - Constant Step Size}. In the following, we present the generalization error of ZoSS for an unbounded loss with a decreasing step size. Recall that the results for unbounded nonconvex loss also hold for the case of a convex loss with similar bounds on the generalization error and learning rate (see the first case of Theorem \ref{thm:Nonconvex Convex Unbounded Loss case2 - Constant Step Size}). 
\begin{theorem}[\textbf{Unbounded Loss | Decreasing Step Size}]\label{thm:Nonconvex Convex Unbounded Loss - Decreasing Step Size}
Assume that the loss $f(\cdot, z)$ is $L$-Lipschitz, $\beta$-smooth for all $z\in \mc{Z}$. Consider ZoSS with update rule \eqref{eq:ZO_updates}, $T$ the total number of iterates, $\alpha_t \leq C/ t \MDK$ for all $t\leq T$ and for some $C>0$, and $\mu\leq cL\MDK/(n\beta (3+d)^{3/2}) $ for some $c>0$. Then the generalization error of ZoSS is bounded by
\begin{align}
  \!\!\!\!  \eps_{\mathrm{gen}} \leq \frac{   (2+c)L^2 (eT)^{C\beta} }{n   } \min\left\{ C + \beta^{-1}, C\log (eT) \right\}.
\end{align}
\end{theorem} For the proof of Theorem \ref{thm:Nonconvex Convex Unbounded Loss - Decreasing Step Size} see Appendix \ref{App_Proof_thm:Nonconvex Convex Unbounded Loss - Decreasing Step Size}. Note that the constant $C$ is free and controls the learning rate. Furthermore, it quantifies the trade-off between the speed of training and the generalization of the algorithm.

In the next section, we consider the ZoSS algorithm with a mini-batch of size $m$ for which we provide generalization error bounds. These results hold under the assumption of unbounded loss and for any batch size $m$ including the case $m=1$.  

\section{Generalization of Mini-Batch ZoSS}\label{Section mini_batch}
For the \textit{mini-batch} version of ZoSS, at each iteration $t$, the randomized selection rule (uniformly) samples a batch $J_t$ of size $m$ and evaluates the new direction of the update by averaging the smoothed approximation ${\Delta f}^{K,\mu}_{w,z}$ over the samples $z\in J_t$ as  \begin{align}
  {\Delta f}^{K,\mu}_{w,J_{t}} &\equiv 
  {\Delta f}^{K,\mu,\mathbf{U}^t}_{w,J_{t}}\triangleq \frac{1}{mK}\sum^{m}_{i=1} \sum^K_{k=1}\frac{f(w +\mu U^t_{k,i},z_{J_{t,i}})-f(w ,z_{J_{t,i}} )}{\mu}U^t_{k,i},
\end{align} where $U^t_{k,i} \sim \mc{N}(0 ,I_d)$ are i.i.d. (standard normal), and $\mu\in \mbb{R}^+$. The update rule of the mini-batch ZoSS is $W_{t+1} = W_{t} -\alpha_t {\Delta f}^{K,\mu}_{W_t,J_{t}} $ for all $t\leq T$. Due to space limitation, we refer the reader to Appendix \ref{Appendix_ZO_MB} for the detailed stability analysis of ZoSS with mini-batch. Specifically, we prove a growth recursion lemma for the mini-batch ZoSS updates (Appendix \ref{Results_ZO_APP}, Lemma \ref{lemma:ZoSS Growth Recursion mb}). Although the iterate stability error in the growth recursion depends on the batch size $m$, the stability bound on the final iterates is independent of $m$, and coincides with the single example updates ($m=1$, Lemma \ref{thm:recur_solution_nonconvex}). Herein, we provide an informal statement of the result. 
\begin{lemma}[Mini-Batch ZoSS Stability | Nonconvex Loss]\label{thm_informalZossStability}
Consider the mini-batch ZoSS with any batch size $m\leq n$, and iterates $W_{t+1} = W_{t} -\alpha_t {\Delta f}^{K,\mu}_{W_t,J_{t}}  $, $W'_{t+1} = W'_{t} -\alpha_t {\Delta f}^{K,\mu}_{W'_t ,J'_{t}}  $, for all $t\leq T$, 
with respect to the sequences $S,S'$. Then the stability error $\delta_T$ satisfies the inequality of Lemma \ref{thm:recur_solution_nonconvex}.
\end{lemma} We refer the reader to Appendix Section \ref{Results_ZO_APP}, Theorem \ref{thm:recur_solution_nonconvex_mb} for the formal statement of the result.\footnote{As in the single-query ($m=1$) ZoSS, under the assumption of convex loss, the stability error of mini-batch ZoSS satisfies the inequality \eqref{eq:convex_stability_appendix}, Appendix \ref{Appendix_Complementary_results}, Lemma \ref{thm:recur_solution_convex}.} Through the Lipschitz condition of the loss and Lemma \ref{thm_informalZossStability}, we show that the mini-batch ZoSS enjoys the same generalization error bounds as in the case of single-query ZoSS ($m=1$). As a consequence, the batch size does not affect the generalization error.
\begin{theorem}[Mini-batch ZoSS | Unbounded Loss]
Theorem \ref{thm:Nonconvex Convex Unbounded Loss case2 - Constant Step Size} and Theorem \ref{thm:Nonconvex Convex Unbounded Loss - Decreasing Step Size} hold for the mini-batch ZoSS with iterate $W_{t+1} = W_{t} -\alpha_t {\Delta f}^{K,\mu}_{W_t,J_{t}}  $, for all $t\leq T$ and any batch size $m\leq n$.
\end{theorem}
By letting $K\rightarrow \infty $ and $c\rightarrow 0$, the generalization error bounds of mini-batch ZoSS reduce to those of mini-batch SGD, extending results of the single-query ($m=1$) SGD that appeared in prior work~\cite{hardt2016train}. Additionally, once $K\rightarrow \infty $, $c\rightarrow 0$ and $m=n $ we obtain generalization guarantees for full-batch GD. For the sake of clarity and completeness we provide dedicated stability and generalization analysis of full-batch GD in Appendix \ref{Append_GD}, Corollary \ref{corollary_GD}. 

\section{Discussion: Black-box
Adversarial Attack Design and Future Work}
A standard, well-cited example of ZoSS application is adversarial learning as considered in~\cite{chen2017zoo}, when the gradient is not known for the adversary (for additional applications for instance federated/reinforcement learning, linear quadratic regulators; see also Section \ref{intro} for additional references). Notice that the algorithm in~\cite{chen2017zoo} is restrictive in the high dimensional regime since it requires $2d$ function evaluations per iteration. In contrast, ZoSS can be considered with any $K\geq 2$ functions evaluations (the trade-off is between accuracy and resource allocation, which is also controlled through $K$). If $K=d+1$ evaluations are available we recover guarantees for the deterministic zeroth-order approaches (similar to~\cite{chen2017zoo}). 

Retrieving a large number of function evaluations often is not possible in practice. When a limited amount of function evaluations is available, the adversary obtains the solution (optimal attack) with an optimization error that scales by a factor of $\sqrt{d/K}$, and the generalization error of the attack is of the order $\sqrt{T}/n$ under appropriate choices of the step-size, the smoothing parameter $\mu$ and $K$. Fine tuning of the these parameters might be useful in practice, but in general $K$ should be chosen as large as possible. In contrast, $\mu$ should be small and satisfy the inequality $\mu\leq   cL\MDK/n \beta (3+d)^{3/2} $ (Theorem \ref{thm:Nonconvex Convex Unbounded Loss - Decreasing Step Size}). For instance, in practice $\mu$ is often chosen between $10^{-10}$ and $10^{-8}$ (or even lower) and the ZoSS algorithm remains (numerically) stable.

For neural networks with smooth activation functions~\cite{hendrycks2016gaussian,ramachandran2017searching,7280459}, the ZoSS algorithm does not require the smoothness parameter $\beta$ to be necessarily known, however if $\beta$ is large then the guarantees of the estimated model would be pessimistic. To ensure that the learning procedure is successful, the adversary can approximate $\beta$ (since the loss is not known) by estimating the (largest eigenvalue of the) Hessian through the available function evaluations~\cite[Section 4.1]{balasubramanian2022zeroth}. 

Although the non-smooth (convex) loss setting lies out of the scope of this work, it is expected to inherit properties and rates of the SGD for non-smooth losses (at least for sufficiently small smoothing parameter $\mu$). In fact,~\cite[page 3, Table 1]{bassily2020stability} developed upper and lower bounds for the SGD in the non-smooth case, and they showed that standard step-size choices provide vacuous stability bound. Due to these inherent issues of non-smooth (and often convex only cases), the generalization error analysis of ZoSS for non-smooth losses remains open. Finally, information-theoretic generalization error bounds of ZoSS can potentially provide further insight into the problem, due to the noisy updates of the algorithm, and consist part of future work.

\section{Conclusion}
In this paper, we characterized the generalization ability of black-box learning models. Specifically, we considered the Zeroth-order Stochastic Search (ZoSS) algorithm, which evaluates smoothed approximations of the unknown gradient of the loss by only relying on $K+1$ loss evaluations. Under the assumptions of a Lipschitz and smooth (unknown) loss, we showed that the ZoSS algorithm achieves the same generalization error bounds as that of SGD, while the learning rate is slightly decreased compared to that of SGD. The efficient generalization ability of ZoSS, together with strong optimality results related to the optimization error by Duchi et al.~\cite{duchi2015optimal}, makes it a robust and powerful algorithm for a variety of black-box learning applications and problems.

%

\bibliographystyle{unsrturl}
\bibliography{bibliography.bib}

\appendices
\section{Proofs}\label{Appendix_A}

\subsection{Proof of Lemma \ref{lemma:variance_reduction_zero_mean}}\label{App_Proof_Variance_reduction}
 
 For fixed ${\bf V}\in\mbb{R}^d$, we have due to independence
 \begin{align}
\mathbb{E}\Bigg[\Bigg\Vert\dfrac{1}{K}\sum_{k=1}^{K}\langle{\bf V},{\bf U}_{k}\rangle{\bf U}_{k}-{\bf V}\Bigg\Vert^{2}\Bigg] & =\dfrac{1}{K^{2}}\mathbb{E}\Bigg[\Bigg\Vert\sum_{k=1}^{K}\langle{\bf V},{\bf U}_{k}\rangle{\bf U}_{k}-{\bf V}\Bigg\Vert^{2}\Bigg]\nonumber \\
 & =\dfrac{1}{K^{2}}\sum_{k=1}^{K}\mathbb{E}\big[\Vert\langle{\bf V},{\bf U}_{k}\rangle{\bf U}_{k}-{\bf V}\Vert^{2}\big]\nonumber \\
 & =\dfrac{1}{K}\mathbb{E}\big[\Vert\langle{\bf V},{\bf U}_{1}\rangle{\bf U}_{1}-{\bf V}\Vert^{2}\big]\nonumber.
\end{align}
Now, again due to independence
\begin{align}
\mathbb{E}\big[\Vert\langle{\bf V},{\bf U}_{1}\rangle{\bf U}_{1}-{\bf V}\Vert^{2}\big] & =\mathbb{E}\big[\Vert\langle{\bf V},{\bf U}_{1}\rangle{\bf U}_{1}\Vert^{2}-2\langle\langle{\bf V},{\bf U}_{1}\rangle{\bf U}_{1},{\bf V}\rangle+\Vert{\bf V}\Vert^{2}\big]\nonumber \\
 & =\mathbb{E}\big[\big(\langle{\bf V},{\bf U}_{1}\rangle\big)^{2}\Vert{\bf U}_{1}\Vert^{2}\big]-2\mathbb{E}\big[\langle{\bf V},{\bf U}_{1}\rangle\langle{\bf U}_{1},{\bf V}\rangle\big]+\Vert{\bf V}\Vert^{2}\nonumber \\
 & ={\bf V}^{T}\mathbb{E}\big[{\bf U}_{1}{\bf U}_{1}^{T}\Vert{\bf U}_{1}\Vert^{2}\big]{\bf V}-2{\bf V}^{T}\mathbb{E}\big[{\bf U}_{1}{\bf U}_{1}^{T}\big]{\bf V}+\Vert{\bf V}\Vert^{2}\nonumber \\
 & =\sum_{i=1}^{d}{\bf V}^{T}\mathbb{E}\big[{\bf U}_{1}{\bf U}_{1}^{T}\big(U_{1}^{(i)}\big)^{2}\big]{\bf V}-2\Vert{\bf V}\Vert^{2}+\Vert{\bf V}\Vert^{2}\nonumber \\
 & \le\sum_{i=1}^{d}3{\bf V}^{T}{\bf V}-\Vert{\bf V}\Vert^{2}\nonumber \\
 & =(3d-1)\Vert{\bf V}\Vert^{2}\nonumber.
\end{align}
Therefore,
\begin{equation}\nonumber
\mathbb{E}\bigg[\bigg\Vert\dfrac{1}{K}\sum_{k=1}^{K}\langle{\bf V},{\bf U}_{k}\rangle{\bf U}_{k}-{\bf V}\bigg\Vert^{2}\bigg]\le\dfrac{(3d-1)\Vert{\bf V}\Vert^{2}}{K}.
\end{equation}
Thus, if ${\bf V}$ is random and independent of all ${\bf U}_k$'s, it follows that
%
%
\begin{align*}
      \mathbb{E}\Bigg[\Bigg\Vert\dfrac{1}{K}\sum_{k=1}^{K}\langle{\bf V},{\bf U}_{k}\rangle{\bf U}_{k}-{\bf V}\Bigg\Vert\Bigg|  \mathbf{V}\Bigg]   
     &\leq  \sqrt{ \mathbb{E}\Bigg[\Bigg\Vert\dfrac{1}{K}\sum_{k=1}^{K}\langle{\bf V},{\bf U}_{k}\rangle{\bf U}_{k}-{\bf V}\Bigg\Vert^{2}\Bigg|  \mathbf{V}\Bigg] } \\
     &\leq \sqrt{\frac{3d-1}{K} \Vert  \mathbf{V} \Vert ^2} 
     \\
     &=  \sqrt{\frac{3d-1}{K}}  \Vert  \mathbf{V} \Vert, 
\end{align*} and our claim is proved. \qedwhite

\subsection{Proof of Corollary \ref{thm:coroallry_dimensional_free}}\label{Appendix_proof_corollary}
 Denote by $W_0 (\cdot)$ the Lambert function~\cite{corless1996lambertw}. Through Theorem \ref{thm:nonconvex-bounded-loss_tight} and by replacing $C$ with $C\MDK$ to recover the required learning rate, the generalization error is bounded as
\begin{align*}
     &\eps_{\mathrm{gen}} \\&\leq  \frac{1+(\beta C\MDK)^{-1}}{n} ((2+c)CL^2)^{\frac{1}{1+\beta C\MDK}} (\MDK)^{\frac{1}{1+\beta C \MDK}} (eT)^\frac{\beta C\MDK}{\beta C\MDK+1}\\
    & \leq \frac{1+(\beta C \MDK)^{-1}}{n} ((2+c)CL^2)^{\frac{1}{1+\beta C\MDK}} \lp \frac{1}{\beta C W_0 \lp\frac{1}{\beta C e}\rp}\rp^{\frac{1}{1+1/W_0 \lp\frac{1}{\beta C e}\rp}} (eT)^\frac{\beta C  \MDK}{\beta C  \MDK+1}\numberthis \label{eq:max1}\\
    & \leq \frac{1+(\beta C \MDK)^{-1}}{n} ((2+c)CL^2)^{\frac{1}{1+\beta C\MDK}} \lp \frac{1}{ W_0 \lp 1/e \rp}\rp^{\frac{1}{1+1/W_0 \lp\frac{1}{ e}\rp}}  \max\{1, (\beta C)^{-1} \} (eT)^\frac{\beta C  \MDK}{\beta C  \MDK+1}\\
    & \leq \frac{3}{2} \frac{1+(\beta C \MDK)^{-1}}{n} ((2+c)CL^2)^{\frac{1}{1+\beta C\MDK}}    \max\{1, (\beta C)^{-1} \} (eT)^\frac{\beta C  \MDK}{\beta C  \MDK+1}\numberthis \label{eq:max2}\\
    & \leq \frac{3}{2} \frac{1+(\beta C )^{-1}}{n} ((2+c)CL^2)^{\frac{1}{1+\beta C\MDK}}    \max\{1, (\beta C)^{-1} \} (eT)^\frac{\beta C  \MDK}{\beta C  \MDK+1}\numberthis \label{eq:max3}\\
    & \leq \frac{3}{2} \frac{1+(\beta C )^{-1}}{n} \max\{1,(2+c)CL^2\}    \max\{1, (\beta C)^{-1} \} (eT)^\frac{\beta C  \MDK}{\beta C  \MDK+1}\numberthis\label{eq:max4}\\
    &\leq \frac{3}{2} \frac{\lp 1+(\beta C )^{-1}\rp^2}{n} \lp 1 + (2+c)CL^2\rp    (eT)^\frac{\beta C  \MDK}{\beta C  \MDK+1}\\
    &\leq \lp 1+(\beta C )^{-1}\rp^2 \lp 1 + (2+c)CL^2\rp \frac{3Te}{2n},\numberthis \label{eq:max5}
\end{align*} the maximization of $x^{1/(1+x\beta C )}$ gives \eqref{eq:max1}, we find \eqref{eq:max2} by maximizing the $\lp  \beta C \rp^{-1/(1+1/W_0 \lp 1/xe\rp)} $, and $  W_0 \lp 1/ x e\rp^{-1 / ( 1+1/W_0 \lp 1/ x e\rp)}$, and by applying the inequality $  W_0 \lp 1/  e\rp^{-1 / ( 1+1/W_0 \lp 1/ x e\rp)}\leq 3/2$. Inequality \eqref{eq:max3} holds since $\MDK\geq 1$, we find \eqref{eq:max4} by maximizing the function $((2+c)CL^2)^{1/(1+\beta C x)}$ for both cases $((2+c)CL^2)<1$ and $((2+c)CL^2)\geq 1$. Finally, \eqref{eq:max5} holds for any value of $d\in\mbb{N}$ and $K\in\mbb{N}$ and gives the bound of the corollary. \qedwhite

\subsection{Proof of Theorem \ref{thm:Nonconvex Convex Unbounded Loss case2 - Constant Step Size}}\label{Appendix_proof_thm:Nonconvex Convex Unbounded Loss case2 - Constant Step Size} We start by proving the first case of the Theorem for both convex and nonconvex loss.
\paragraph{Proof of Theorem \ref{thm:Nonconvex Convex Unbounded Loss case2 - Constant Step Size}, First Case:} Let $\mc{C}$ denote the set of convex loss functions. Under the assumption $\alpha_t \leq C'/T $, and $\mu\leq cL\MDK/(n\beta (3+d)^{3/2}) $ Lemma \ref{thm:recur_solution_nonconvex} (nonconvex loss) and Lemma \ref{thm:recur_solution_convex} (convex loss) give
\begin{align*}
   \!\! &\E[\delta_T | \mc{E}_{\delta_0}]\nonumber \\
   &\leq \frac{  \lp 2     +   c \rp L\MDK  }{n}   \sum^T_{t=1} \frac{C'}{T} \prod^T_{j=t+1} \lp 1 + \frac{\beta C'}{T} \lp \isnonconvex+\sqrt{\frac{3d-1}{K}}\rp\rp\\
   &=\frac{  \lp 2     +   c \rp L C'\MDK }{T n}  \sum^T_{t=1}   \lp 1 + \frac{\beta C'}{T} \lp \isnonconvex+\sqrt{\frac{3d-1}{K}}\rp\rp^{T-t}\\
   &= \frac{  \lp 2     +   c \rp L C'\MDK }{Tn}\lp 1 + \frac{\beta C'}{T} \lp \isnonconvex+\sqrt{\frac{3d-1}{K}}\rp\rp^{T}  \sum^T_{t=1}   \lp 1 + \frac{\beta C'}{T} \lp \isnonconvex+\sqrt{\frac{3d-1}{K}}\rp\rp^{-t}\\
   & = \frac{  \lp 2     +   c \rp L C'\MDK }{Tn}\frac{\lp 1 + \frac{\beta C'}{T} \lp \isnonconvex+\sqrt{\frac{3d-1}{K}}\rp\rp^{T}-1}{\lp 1 + \beta \frac{C'}{T} \lp \isnonconvex+\sqrt{\frac{3d-1}{K}}\rp\rp-1}   \\
   &= \frac{  \lp 2     +   c \rp L\MDK  }{n}\frac{\lp 1 + \frac{\beta C'}{T} \lp \isnonconvex+\sqrt{\frac{3d-1}{K}}\rp\rp^{T}-1}{ \beta  \lp \isnonconvex+\sqrt{\frac{3d-1}{K}}\rp} \\
   &\leq  \frac{  \lp 2     +   c \rp L\MDK  }{n}\frac{ \exp \lp  \beta C' \lp \isnonconvex+\sqrt{\frac{3d-1}{K}}\rp \rp  -1}{ \beta  \lp \isnonconvex+\sqrt{\frac{3d-1}{K}}\rp}.
\end{align*} If the loss is convex ($f(\cdot)\in \mc{C}$) and $\alpha_t \leq C'/T\leq 2/\beta$, the last display under the choice \begin{align}
    C'=\frac{\log(1+C\beta\sqrt{\frac{3d-1}{K}}/\MDK)}{\beta \sqrt{\frac{3d-1}{K}}}
\end{align} gives \begin{align}
    \E[\delta_T | \mc{E}_{\delta_0}] &\leq \frac{  \lp 2     +   c \rp L\MDK  }{n}\frac{ \exp \lp  \beta C' \sqrt{\frac{3d-1}{K}}\rp  -1}{ \beta  \sqrt{\frac{3d-1}{K}}}\leq \frac{  \lp 2     +   c \rp C L  }{n}.
\end{align} If the loss is nonconvex ($f(\cdot)\notin \mc{C}$) and $\alpha_t \leq C''/T$, then by choosing \begin{align}
    C'' =  \frac{\log(1+C\beta)}{\beta \MDK},
\end{align} we find \begin{align}
    \E[\delta_T | \mc{E}_{\delta_0}] &\leq \frac{  \lp 2     +   c \rp L \MDK }{n}\frac{ \exp \lp  \beta C''\MDK \rp  -1}{ \beta  \MDK }\leq \frac{  \lp 2     +   c \rp C L  }{n}.
\end{align}
The Lipschitz assumption $\E \left[ |f(W_T,z) - f(W'_T,z)| \right]\leq L  \E[\delta_T] = L  \E[\delta_T | \mc{E}_{\delta_0}]]$ (as a consequence of $\P (\mc{I}\leq 0)= \P (\delta_0 > 0)=0$) completes the proof for the first case of the theorem.

\paragraph{Proof of Theorem \ref{thm:Nonconvex Convex Unbounded Loss case2 - Constant Step Size}, Second Case:}
%
Lemma \ref{thm:recur_solution_nonconvex} (nonconvex loss) under the choice $t_0 = 0$ gives 
\begin{align*}
   \!\! &\E[\delta_T | \mc{E}_{\delta_0}]\nonumber 
\leq\lp \frac{2 L}{n} \MDK    +   \mu \beta (3+d)^{3/2} \rp  \sum^T_{t=1} \alpha_t \prod^T_{j=t+1} \lp 1 + \beta \alpha_j \lp 1+\sqrt{\frac{3d-1}{K}}\rp\rp.\numberthis
\end{align*} Under the assumption $\alpha_t \leq C/( T \MDK)$, and $\mu\leq cL\MDK/(n\beta (3+d)^{3/2}) $ we find
\begin{align*}
   \!\! &\E[\delta_T | \mc{E}_{\delta_0}]\nonumber \\
   &\leq \lp 2     +   c \rp L\frac{  \MDK  }{n}   \sum^T_{t=1} \frac{C}{ T \MDK} \prod^T_{j=t+1} \lp 1 + \frac{C\beta}{T}\rp\\
   &=\lp 2     +   c \rp L\frac{  \MDK  }{n}  \frac{C}{ T \MDK} \sum^T_{t=1}   \lp 1 + \frac{C\beta}{T}\rp^{T-t}\\
   &= \lp 2     +   c \rp L\frac{  \MDK  }{n}  \frac{C \lp 1 + \frac{C\beta}{T}\rp^{T}}{ T \MDK} \sum^T_{t=1}   \lp 1 + \frac{C\beta}{T}\rp^{-t}\\
   &= \frac{  \lp 2     +   c \rp L  }{n}  \frac{C }{ T } \frac{\lp 1 + \frac{C\beta}{T}\rp^{T}-1}{\lp 1 + \frac{C\beta}{T}\rp-1}\\
   &= \frac{  \lp 2     +   c \rp L  }{n}  \frac{\lp 1 + \frac{C\beta}{T}\rp^{T}-1}{\beta}.\numberthis
\end{align*}
The Lipschitz assumption $\E \left[ |f(W_T,z) - f(W'_T,z)| \right]\leq L  \E[\delta_T] = L  \E[\delta_T | \mc{E}_{\delta_0}]]$ (as a consequence of $\P (\mc{I}\leq 0)= \P (\delta_0 > 0)=0$) completes the proof. \qedwhite

\subsection{Proof of Theorem \ref{thm:Nonconvex Convex Unbounded Loss - Decreasing Step Size}}\label{App_Proof_thm:Nonconvex Convex Unbounded Loss - Decreasing Step Size} Under the assumptions $\mu\leq cL\MDK/(n\beta (3+d)^{3/2}) $ and $\alpha_t \leq C/ t \MDK$, Lemma \ref{thm:recur_solution_nonconvex} gives \begin{align*}
   \!\! &\E[\delta_T | \mc{E}_{\delta_0}]\nonumber 
   \\
   &\leq \lp \frac{2 L}{n} \MDK    +   \mu \beta (3+d)^{3/2} \rp  \sum^T_{t=1} \alpha_t \prod^T_{j=t+1} \lp 1 + \beta \alpha_j \lp 1+\sqrt{\frac{3d-1}{K}}\rp\rp\\
   &\leq  \frac{  \MDK  }{n} (2+c)L \sum^T_{t=1} \frac{C}{ t \MDK} \prod^T_{j=t+1} \lp 1 + \frac{C\beta}{j}\rp\\
   & \leq  \frac{   (2+c)L  }{n} \sum^T_{t=1} \frac{C}{ t } \exp \lp  \sum ^T_{j=t+1} \frac{C\beta}{j} \rp\\
   &\leq  \frac{   (2+c)L  }{n}  \sum^T_{t=1} \frac{C}{ t } \exp \lp C\beta \log \lp  \frac{eT}{t+1}\rp \rp\\
   & =  \frac{  C (2+c)L}{n   }  \sum^T_{t=1} \frac{1}{t} \lp  \frac{eT}{t+1}\rp^{C\beta} \\
   &\leq \frac{  C (eT)^{C\beta} (2+c)L }{n   }  \sum^T_{t=1} \frac{1}{t^{C\beta+1}}\\
   &\leq \frac{   (eT)^{C\beta} (2+c)L}{n   } \min\left\{ \frac{C\beta+1}{\beta}, C\log (e T) \right\}.\numberthis
\end{align*} The last inequality holds because of the inequalities $\sum^T_{t=1} t^{-C\beta-1}\leq \sum^\infty_{t=1} t^{-C\beta-1}\!\leq \! (C\beta +1)/C\beta$ and $\sum^T_{t=1} t^{-C\beta-1}\leq \sum^T_{t=1} 1/t\leq \log(e T) $. Then the inequality $\E \left[ |f(W_T,z) - f(W'_T,z)| \right]\leq L  \E[\delta_T] = L  \E[\delta_T | \mc{E}_{\delta_0}]]$ (as a consequence of $\P (\mc{I}\leq 0)= \P (\delta_0 > 0)=0$) completes the proof. \qedwhite

\section{Complementary Results}\label{Appendix_Complementary_results}
In this Section we provide complementary results and the corresponding proofs. The next result provides the equivalent bound of Lemma \ref{thm:recur_solution_nonconvex} for convex losses. 
\begin{lemma}[ZoSS Stability \textbf{Convex Loss}]\label{thm:recur_solution_convex}
Let the loss function $f(\cdot, z)$ be $L$-Lipschitz, convex and $\beta$-smooth for all $z\in \mc{Z}$. Consider the ZO-SM algorithm \eqref{eq:ZO_updates} with parameters estimates $W_T$ and $W'_T$ for all the data-sets $S,S'$ respectively (that differ in exactly one entry). Then the discrepancy $\delta_T \triangleq \Vert  W_{T} - W'_{T} \Vert$ under the event $\mc{E}_{\delta_{t_0}}$ satisfies the following inequality,
\begin{align}\label{eq:convex_stability_appendix}
    &\E[\delta_T|\mc{E}_{\delta_{t_0}}]\leq \lp \frac{2 L}{n} \MDK    +   \mu \beta (3+d)^{3/2} \rp  \sum^T_{t=t_0 +1} \alpha_t \prod^T_{j=t+1} \lp 1 + \beta \alpha_j \sqrt{\frac{3d-1}{K}} \rp.
\end{align}
\end{lemma} We prove Lemma \ref{thm:recur_solution_convex} in parallel with Lemma \ref{thm:recur_solution_nonconvex} in Section \ref{SubSec_Lemmas} of the main part of the paper.

\section{ZoSS with Mini-Batch (Section \ref{Section mini_batch})}\label{Appendix_ZO_MB}
For the stability analysis of mini-batch ZoSS, we similarly consider the sequences $S,S'$ that differ in one example. At each time $t$ we sample a batch $J_t \subset S$ ($J'_t \subset S'$) (and batch size $m\triangleq |J_t|=|J'_t|$) with replacement, or by considering random permutation of the samples and then sample the first $m$ examples. As a consequence in both cases $\P (J_t \neq J'_t) = m/n $. Under the event $\{ J_t \neq J'_t \}$, the sets $J_t , J'_t$ differ in one example $z_{i^*}\neq z'_{i^*}$ (for some $i^*$ without loss of generality), and $z_{i}= z'_{i}$ for any $z_i\in J_{t}$ and $i\in\{1,\ldots,m\}\setminus\{i^*\}$. Let $U^t_{k,i}\sim \mc{N}(0 ,I_d)$ be independent for all $k\in\{1,2,\ldots,K\}$, $i\in\{1,2,\ldots,m\}$ and $t\leq T$ and $\mu\in \mbb{R}^+$. Recall the definition of the smoothed approximation and update rule mapping of mini-batch ZoSS, \begin{align}
  {\Delta f}^{K,\mu}_{w,J_{t}} &\equiv 
  {\Delta f}^{K,\mu,\mathbf{U}^t}_{w,J_{t}}\triangleq \frac{1}{mK}\sum^{m}_{i=1} \sum^K_{k=1}\frac{f(w +\mu U^t_{k,i},z_{J_{t,i}})-f(w ,z_{J_{t,i}} )}{\mu}U^t_{k,i},\\
  \tilde{G}_{\!J_t}(w) &\triangleq w - \alpha_t {\Delta f}^{K,\mu}_{w,J_{t}} , \quad \tilde{G}'_{\!J'_t}(w) \triangleq w - \alpha_t {\Delta f}^{K,\mu}_{w,J'_{t}} . \label{eq:ZO_updates_mini_batch}
\end{align} For the stability error decomposition, we define the gradient based mappings $G_{\!J_t}(\cdot)$ and $G'_{\!J'_t}(\cdot)$ as \begin{align*}
    G_{\!J_t}(w) \triangleq w - \alpha_t \nabla f_{w , J_t },\,\,\,
    G'_{\!J'_t}(w) \triangleq w - \alpha_t\nabla f_{w , J'_t } ,\text{ and }\nabla f_{w , J_t }\triangleq \frac{1}{m} \sum_{z \in J_t} \nabla f(w ,z ) \numberthis
\end{align*} the iterate stability error of the mini-batch ZoSS $ \tilde{G}_{\!J_t}(w) -  \tilde{G}'_{\!J'_t}(w')$ at time $t$  and similarly to \eqref{eq:error_decomposition}, we show that \begin{align*}
& \tilde{G}_{\!J_t}(w) -  \tilde{G}'_{\!J'_t}(w')\propto \underbrace{  G_{\!J_t}(w) -   G'_{\!J'_t}(w') }_{\eps^m_{\text{GBstab}}} + \underbrace{\big[ \nabla f_{w , J_t } - {\Delta f}^{K,\mu}_{w,J_{t}} \big] + \big[\nabla f_{w' , J'_t } - {\Delta f}^{K,\mu}_{w',J'_{t}} \big]}_{\eps^m_{\text{est}}}.\numberthis\label{eq:error_decomposition_MB}
\end{align*} For the mini-batch case the derivation of the stability differs to that of \eqref{eq:error_decomposition} ($m=1$).To analyze the error term $\eps^m_{\text{GBstab}}$ in the mini batch case, we derive (for the proof see Appendix, Section \ref{GR_RECUR_SGD_PROOF}) and apply the Mini-Batch SGD Growth Recursion for the mappings $G_{J_t}(\cdot),G'_{J'_t}(\cdot)$, that is an extension of~\cite[Lemma 2.4]{hardt2016train} and describes the growth recursion property of the SGD algorithm with mini batch. 
\begin{lemma}[Mini-Batch SGD Growth Recursion]\label{lemma:Mini-Batch SGD Growth Recursion}
Let $\{G_{\!J_t}\}^T_{t=1}$ and $\{G'_{\!J'_t}\}^T_{t=1}$ be arbitrary sequences of updates. Let $w_0 =w'_0$ be the starting point, $w_{t+1} = G_{\!J_t} (w_t) $ and $w'_{t+1} = G'_{\!J'_t} (w'_t) $ for any $t\in \{1,\ldots , T\}$. Then for any $t\geq 0$ the following recursion holds
\begin{align}
   \Vert G_{\!J_t} (w_t) - G'_{\!J'_t} (w'_t) \Vert \leq  
     \begin{cases}
       (1 + \beta\alpha_t )\Vert w_t - w'_t \Vert &\text{ if }G_{\!J_t}(\cdot) =G'_{\!J'_t}(\cdot) \\
       \lp 1+\frac{m-1}{m} \beta \alpha_t \rp  \Vert w_t - w'_t \Vert + \frac{2}{m} L \alpha_t &\text{ if } G_{\!J_t}(\cdot) \neq G'_{\!J'_t}(\cdot).
     \end{cases}
\end{align}
\end{lemma} The error $\eps^m_{\text{GBstab}}$ depends on the batch size $m$ at the event of different batch selection $\{ J_t \neq J'_t\}$ as appears in Lemma \ref{lemma:Mini-Batch SGD Growth Recursion}. Additionally, the error $\eps^m_{\text{est}}$ breaks down into the errors $\eps_{\mu},\eps^m_{d/K}$. Although $\eps_{\mu}$ is independent of $m$, $\eps_{\mu}\leq \mu \beta \E[\Vert U \Vert^3 ]$ ($U\in\mbb{R}^d$ is standard normal), $\eps^m_{\text{est}}$ dependents on the batch size $m$ similarly to gradient based stability error $\eps^m_{\text{GBstab}}$. If the randomized algorithm (at time $t$) selects $J_t\neq J'_t$ then $\eps^m_{d/K}\leq 2\beta\alpha_t \sqrt{d/K}[(m-1)\Vert w-w'\Vert +2L]/m $, else $\eps^{m}_{d/K}\leq  4\beta\alpha_t \Vert w-w' \Vert \sqrt{d/K}$. We provide a unified representation of the stability error $\tilde{G}_{\!J_t}(w) -  \tilde{G}'_{\!J'_t}(w')$ in the Section (Mini-Batch ZoSS Growth Recursion, Lemma \ref{lemma:ZoSS Growth Recursion mb}). 

\subsection{Results for the ZoSS with Mini-Batch}\label{Results_ZO_APP}
We start by providing the growth recursion lemma for the ZoSS with mini batch.
\begin{lemma}[Mini-Batch ZoSS Growth Recursion]\label{lemma:ZoSS Growth Recursion mb}
Consider the sequences $\{\tilde{G}_{\!J_t}\}^T_{t=1}$ and $\{\tilde{G}'_{\!J_t}\}^T_{t=1}$ and $\mu\leq cL\MDK/(n\beta (3+d)^{3/2}) $. Let $w_0 =w'_0$ be the starting point, $w_{t+1} = \tilde{G}_{\!J_t} (w_t) $ and $w'_{t+1} = \tilde{G}'_{\!J_t} (w'_t) $ for any $t\in \{1,\ldots , T\}$. Then for any $w_t,w'_t\in\mbb{R}^d$ and $t\geq 0$ the following recursion holds
\begin{align*}
   \E [\Vert \tilde{G}_{\!J_t} (w_t) - \tilde{G}'_{\!J_t} (w'_t) \Vert] \leq  
     \begin{cases}
       \lp 1+ \beta\alpha_t\MDK \rp \delta_t   +\frac{cL\alpha_t}{n}\MDK &\text{ if }\tilde{G}_{\!J_t}(\cdot) =\tilde{G}'_{\!J_t}(\cdot) \\
       \lp 1+\frac{m-1}{m} \beta \alpha_t\MDK \rp  \delta_t  + \frac{2L \alpha_t}{m} \MDK  +  \frac{cL\alpha_t}{n}\MDK &\text{ if } \tilde{G}_{\!J_t}(\cdot) \neq \tilde{G}'_{\!J_t}(\cdot).
     \end{cases}
\end{align*}
\end{lemma}

Our next result provides a stability guarantee on the difference of the mini-batch ZoSS outputs $W_T, W'_T$, that holds for any batch size $m$. 
\begin{theorem}[\textbf{Stability of ZoSS with Mini Batch | Nonconvex Loss}]\label{thm:recur_solution_nonconvex_mb}
Assume that the loss function $f(\cdot, z)$ is $L$-Lipschitz and $\beta$-smooth for all $z\in \mc{Z}$. Consider the ZoSS with mini batch of size $m\in\{1,\ldots,n\}$, initial state $W_0 =W'_0$, iterates $W_t = \tilde{G}_{\!J_t}(W_t) $, $W'_t = \tilde{G}'_{\!J'_t}(W'_t) $ for $t>0$, and with final-iterate estimates $W_T$ and $W'_T$ corresponding to the data-sets $S,S'$, respectively (that differ in exactly one entry). Then the discrepancy $\delta_T \triangleq \Vert  W_{T} - W'_{T} \Vert$ under the event $\mc{E}_{\delta_{t_0}}$ and the choice $\mu\leq cL\MDK/(n\beta (3+d)^{3/2}) $ satisfies the inequality
\begin{align}
   \!\! &\E[\delta_T|\mc{E}_{\delta_{t_0}}]\leq \frac{(2+c)L\MDK}{n}  \sum^T_{t=t_0 +1} \alpha_t \prod^T_{j=t+1} \lp 1 + \beta \alpha_j \MDK\lp 1-\frac{1}{n}\rp\rp.\label{eq:delta_bounds_mb}
\end{align}
\end{theorem} We prove Theorem \ref{thm:recur_solution_nonconvex_mb} in Appendix, Section \ref{Appendix_Proof_recur_solution_nonconvex_mb}. Note that\footnote{Under the random selection rule $\P (\mc{I}\leq t_0) = 1- \P (\cap^{t_0}_{i=1} \{\mc{I}\neq i\})= 1-\prod^{t_0}_{i=1} \P ( \{\mc{I}\neq i\}) $.} $\P (\mc{I}\leq t_0)= 1- (1- m/n)^{t_0}$. By setting the free parameter $t_0 = 0 $ and through the Lipschitz assumption we find the stability bound of the loss as $\E \left[ |f(W_T,z) - f(W'_T,z)| \right]\leq L  \E[\delta_T] = L  \E[\delta_T | \mc{E}_{\delta_0}]] $. The last inequality and the solution of the recursion in \eqref{eq:delta_bounds_mb} show that Theorem \ref{thm:Nonconvex Convex Unbounded Loss case2 - Constant Step Size} and Theorem \ref{thm:Nonconvex Convex Unbounded Loss - Decreasing Step Size} hold for the ZoSS algorithm with mini batch, and any batch size $m\in \{1,\ldots,n \}$ as well.

\subsection{Proof of Lemma \ref{lemma:Mini-Batch SGD Growth Recursion}}\label{GR_RECUR_SGD_PROOF}
\noindent Under the assumption of nonconvex losses we find the first part of the statement as
\begin{align*}
    \Vert G_{\!J_t}(w_t) - G_{\!J_t}(w'_t) \Vert &\leq \Vert w_t - w'_t \Vert + \frac{\alpha_t}{m} \bigg\Vert \sum_{z \in J_t} \nabla_w f(w ,z )|_{w=w_t} -  \sum_{z \in J_t} \nabla_w f(w ,z )|_{w=w'_t} \bigg\Vert\\
    &\leq \Vert w_t - w'_t \Vert + \frac{\alpha_t}{m} \sum_{z \in J_t}\bigg\Vert \nabla_w f(w ,z )|_{w=w_t} -   \nabla_w f(w ,z )|_{w=w'_t} \bigg\Vert\\
    &\leq  \Vert w_t - w'_t \Vert + \frac{\alpha_t}{m} \sum_{z \in J_t} \beta \Vert w_t - w'_t \Vert\\
    & = (1 + \beta\alpha_t )\Vert w_t - w'_t \Vert.\numberthis
\end{align*} Further define $J^{-i^*}_t \triangleq J_t\setminus\{ z_{J_{t,i^*}} \}$ and $J'^{-i^*}_t \triangleq J'_t\setminus\{ z'_{J'_{t,i^*}} \}$, and notice that $J^{-i^*}_t = J'^{-i^*}_t$ for any $t\leq T$ w.p. $1$. \begin{align*}
    &\Vert G_{\!J_t}(w_t) - G'_{\!J'_t}(w'_t) \Vert \\&= \bigg\Vert w_t - w'_t - \frac{\alpha_t}{m} \sum_{z \in J_t} \nabla_w f(w ,z )|_{w=w_t} + \frac{\alpha_t}{m} \sum_{z' \in J'_t} \nabla_w f(w ,z' )|_{w=w'_t} \bigg\Vert\\
    &= \bigg\Vert \frac{1}{m} \sum_{z \in J^{-i^*}_t} \lp w_t - \alpha_t \nabla_w f(w ,z )|_{w=w_t}   \rp  - \frac{1}{m} \sum_{z' \in J'^{-i^*}_t} \lp w'_t - \alpha_t \nabla_w f(w ,z' )|_{w=w'_t}   \rp \\
    & \quad + \frac{1}{m} \lp  w_t - \alpha_t \nabla_w f(w ,z_{J_{t,i^*}} )|_{w=w_t} \rp  - \frac{1}{m} \lp w'_t - \alpha_t \nabla_w f(w ,z_{J'_{t,i^*}} )|_{w=w'_t} \rp \bigg\Vert\\
    &= \bigg\Vert \frac{1}{m} \sum_{z \in J^{-i^*}_t} ( \underbrace{w_t - \alpha_t \nabla_w f(w ,z )|_{w=w_t}}_{G(w_t ,z)}  )  - \frac{1}{m} \sum_{z \in J^{-i^*}_t} (\underbrace{ w'_t - \alpha_t \nabla_w f(w ,z )|_{w=w'_t} }_{G(w'_t ,z)}  ) \\
    & \quad + \frac{1}{m} (  \underbrace{w_t - \alpha_t \nabla_w f(w ,z_{J_{t,i^*}} )|_{w=w_t} }_{G(w_t,z_{J_{t,i^*}})} )  - \frac{1}{m} ( \underbrace{w'_t - \alpha_t \nabla_w f(w ,z_{J'_{t,i^*}} )|_{w=w'_t}}_{G'(w'_t,z_{J'_{t,i^*}})} ) \bigg\Vert\\
    &= \frac{1}{m}\bigg\Vert  \sum_{z \in J^{-i^*}_t}\lp G(w_t,z)-G(w'_t,z)\rp  +  G(w_t,z_{J_{t,i^*}})  -  G'(w'_t,z_{J'_{t,i^*}}) \bigg\Vert\\
    &\leq  \frac{1}{m}\bigg\Vert  \sum_{z \in J^{-i^*}_t}\lp G(w_t,z)-G(w'_t,z)\rp\bigg\Vert + \frac{1}{m}\Vert G(w_t,z_{J_{t,i^*}})  -  G'(w'_t,z_{J'_{t,i^*}}) \Vert\\
    &\leq  \frac{1}{m}\sum_{z \in J^{-i^*}_t}\Vert   G(w_t,z)-G(w'_t,z) \Vert +\frac{1}{m} \Vert G(w_t,z_{J_{t,i^*}})  -  G'(w'_t,z_{J'_{t,i^*}}) \Vert. \label{eq:applyLemma2.5}\numberthis
\end{align*}\cite[Lemma 2.4]{hardt2016train} for nonconvex loss ($\eta = 1+\beta \alpha_t$) gives \begin{align}
    \Vert   G(w_t,z)-G(w'_t,z) \Vert &\leq (1+\beta \alpha_t) \delta_t,\\
    \Vert  G(w_t,z_{J_{t,i^*}})-G'(w'_t,z_{J'_{t,i^*}} \Vert &\leq  \delta_t + 2 L \alpha_t.
\end{align} By combining the last two together with \eqref{eq:applyLemma2.5} we find \begin{align*}
    \Vert G_{\!J_t}(w_t) - G'_{\!J'_t}(w'_t) \Vert &\leq \frac{1}{m}\sum_{z \in J^{-i^*}_t}(1+\beta \alpha_t) \delta_t +\frac{1}{m} \lp  \delta_t + 2 L \alpha_t\rp\\
    &= \frac{m-1}{m}(1+\beta \alpha_t) \delta_t +\frac{1}{m} \lp  \delta_t + 2 L \alpha_t\rp\\
    &= \lp 1+\frac{m-1}{m} \beta \alpha_t \rp  \delta_t + \frac{2}{m} L \alpha_t.\numberthis
\end{align*} The last gives the second part of the recursion and completes the proof.\qedwhite
%
%
%
\subsection{Proof of Lemma \ref{lemma:ZoSS Growth Recursion mb} and Theorem \ref{thm:recur_solution_nonconvex_mb}}\label{Appendix_Proof_recur_solution_nonconvex_mb}
First we provide the proof of Lemma \ref{lemma:ZoSS Growth Recursion mb}, then we apply Lemma \ref{lemma:ZoSS Growth Recursion mb} to prove Theorem \ref{thm:recur_solution_nonconvex_mb}.
\paragraph{Proof of Lemma \ref{lemma:ZoSS Growth Recursion mb}} Consider the update rules under the event $\tilde{\mathcal{E}}_t\triangleq \{ \tilde{G}_{\!J_t}(\cdot) \equiv \tilde{G}'_{\!J'_t}(\cdot)\}$ that occurs with probability $\P(\tilde{\mathcal{E}}_t)=1-m/n$ for all $t\leq T$.
Similarly to \eqref{eq:base_diff_1} we find
\begin{align*}
    &\tilde{G}_{\!J_t}(w_t) - \tilde{G}_{\!J_t}(w'_{t})  
    \\& = \underbrace{w_t-\frac{\alpha_t}{m} \sum^m_{i=1} \nabla_{w}f(w ,z_{J_{t,i}})|_{w=w_t}}_{G_{\!J_t}(w_t)} - \Big( \underbrace{w'_t  -\frac{\alpha_t}{m} \sum^m_{i=1}\nabla_{w}f(w ,z_{J_{t,i}})|_{w=w'_t}}_{G_{\!J_t}'(w'_t)\equiv G_{\!J_t}(w'_t)} \Big) \\&\quad - \frac{\alpha_t}{m K}\sum^m_{i=1}\sum^K_{k=1} \lp \frac{\mu}{2} U^{\text{T}}_k \nabla^2_{w}f(w ,z_{J_{t,i}})|_{w=W^*_{k,t,i}} U^t_{k,i} \rp U^t_{k,i}  
    \\
    &\quad+ \frac{\alpha_t}{mK}\sum^m_{i=1}\sum^K_{k=1} \lp \frac{\mu}{2} U^{\text{T}}_k \nabla^2_{w}f(w ,z_{J_{t,i}})|_{w=W^\dagger_{k,t,i}} U^t_{k,i}\rp U^t_{k,i}  \\ &   \quad-   \frac{\alpha_t}{m}\sum^m_{i=1}\bigg( \frac{1}{K}\sum^K_{k=1} \inp{\nabla_{w}f(w ,z_{J_{t,i}})|_{w=w_t}-\nabla_{w}f(w ,z_{J_{t,i}})|_{w=w'_t}}{ U^t_{k,i}}U^t_{k,i} \\&\quad\quad\quad- ( \nabla_{w}f(w ,z_{J_{t,i}})|_{w=w_t}-\nabla_{w}f(w ,z_{J_{t,i}})|_{w=w'_t})\bigg).\numberthis\label{eq:min_batch__diff_1} 
\end{align*} Denote by $\E_{\mathbf{U}^{\otimes K \times m }_t}$ the expectation with respect to product measure of the random vectors $U^t_{k,i}\sim\mc{N}(0,I_d)$ for all $k\in\{1,2,\ldots,K\}$, $i\in\{1,2,\ldots,m\}$ and fixed $t\leq T$. Recall that $U^t_{k,i}$ are independent for all $k\in\{1,2,\ldots,K\}$, $i\in\{1,2,\ldots,m\}$ and $t\leq T$. Inequality \eqref{eq:min_batch__diff_1} and triangle inequality give \begin{align*}
& \E [\Vert\tilde{G}_{\!J_t}(w_t) - \tilde{G}_{\!J_t}(w'_{t}) \Vert ] \\
&\leq  \Vert G_{\!J_t} (w_t) - G_{\!J_t} (w'_t) \Vert  \\&  \quad+ \E_{\mathbf{U}^{\otimes K \times m }_t} \left[ \left\Vert \frac{\alpha_t}{m K}\sum^m_{i=1}\sum^K_{k=1} \lp \frac{\mu}{2} U^{\text{T}}_k \nabla^2_{w}f(w ,z_{J_{t,i}})|_{w=W^*_{k,t,i}} U^t_{k,i} \rp U^t_{k,i} \right\Vert\right] \\& \quad+  \E_{\mathbf{U}^{\otimes K \times m }_t}\left[ \left\Vert\frac{\alpha_t}{mK}\sum^m_{i=1}\sum^K_{k=1} \lp \frac{\mu}{2} U^{\text{T}}_k \nabla^2_{w}f(w ,z_{J_{t,i}})|_{w=W^\dagger_{k,t,i}} U^t_{k,i}\rp U^t_{k,i}\right\Vert\right]\\ & \quad +\E_{\mathbf{U}^{\otimes K \times m }_t}\Bigg[ \Bigg\Vert \frac{\alpha_t}{m}\sum^m_{i=1}\bigg( \frac{1}{K}\sum^K_{k=1} \inp{\nabla_{w}f(w ,z_{J_{t,i}})|_{w=w_t}-\nabla_{w}f(w ,z_{J_{t,i}})|_{w=w'_t}}{ U^t_{k,i}}U^t_{k,i} \\&\quad\quad\quad- ( \nabla_{w}f(w ,z_{J_{t,i}})|_{w=w_t}-\nabla_{w}f(w ,z_{J_{t,i}})|_{w=w'_t})\bigg) \Bigg\Vert\Bigg]\\
\\ &\leq  \Vert G_{\!J_t} (w_t) - G_{\!J_t} (w'_t) \Vert  \\&  \quad+ \E_{\mathbf{U}^{\otimes K \times m }_t} \left[ \left\Vert \frac{\alpha_t}{m K}\sum^m_{i=1}\sum^K_{k=1} \lp \frac{\mu}{2} U^{\text{T}}_k \nabla^2_{w}f(w ,z_{J_{t,i}})|_{w=W^*_{k,t,i}} U^t_{k,i} \rp U^t_{k,i} \right\Vert\right] \\& \quad+  \E_{\mathbf{U}^{\otimes K \times m }_t}\left[ \left\Vert\frac{\alpha_t}{mK}\sum^m_{i=1}\sum^K_{k=1} \lp \frac{\mu}{2} U^{\text{T}}_k \nabla^2_{w}f(w ,z_{J_{t,i}})|_{w=W^\dagger_{k,t,i}} U^t_{k,i}\rp U^t_{k,i}\right\Vert\right]\\ & \quad + \frac{\alpha_t}{m}\sum^m_{i=1}\E_{\mathbf{U}^{\otimes K }_{t,i}}\Bigg\Vert \bigg( \frac{1}{K}\sum^K_{k=1} \inp{\nabla_{w}f(w ,z_{J_{t,i}})|_{w=w_t}-\nabla_{w}f(w ,z_{J_{t,i}})|_{w=w'_t}}{ U^t_{k,i}}U^t_{k,i} \\&\quad\quad\quad- ( \nabla_{w}f(w ,z_{J_{t,i}})|_{w=w_t}-\nabla_{w}f(w ,z_{J_{t,i}})|_{w=w'_t})\bigg) \Bigg\Vert\\
&\leq  \Vert G_{\!J_t} (w_t) - G_{\!J_t} (w'_t) \Vert  \\& \quad + \E_{\mathbf{U}^{\otimes K \times m }_t} \left[ \left\Vert \frac{\alpha_t}{m K}\sum^m_{i=1}\sum^K_{k=1} \lp \frac{\mu}{2} U^{\text{T}}_k \nabla^2_{w}f(w ,z_{J_{t,i}})|_{w=W^*_{k,t,i}} U^t_{k,i} \rp U^t_{k,i} \right\Vert \right] \\&\quad+  \E_{\mathbf{U}^{\otimes K \times m }_t}\left[ \left\Vert\frac{\alpha_t}{mK}\sum^m_{i=1}\sum^K_{k=1} \lp \frac{\mu}{2} U^{\text{T}}_k \nabla^2_{w}f(w ,z_{J_{t,i}})|_{w=W^\dagger_{k,t,i}} U^t_{k,i}\rp U^t_{k,i}\right\Vert\right]\\ & \quad + \frac{\alpha_t}{m}\sum^m_{i=1}  \sqrt{\frac{3d-1}{K}} \Vert \nabla_{w}f(w ,z_{J_{t,i}})|_{w=w_t}-\nabla_{w}f(w ,z_{J_{t,i}})|_{w=w'_t} \Vert \numberthis \label{eq:Lemma1ongradientdifferencesmb}\\
&\leq  \Vert G_{\!J_t} (w_t) - G_{\!J_t} (w'_t) \Vert   + \frac{2\alpha_t }{mK}\sum^m_{i=1}\sum^K_{k=1} \frac{\mu \beta }{2} \E_{\mathbf{U}_{t,i,k}}\left[ \Vert U^t_{k,i}\Vert ^3 \right]  \\&\quad +\frac{\alpha_t}{m}\sum^m_{i=1}  \sqrt{\frac{3d-1}{K}} \Vert \nabla_{w}f(w ,z_{J_{t,i}})|_{w=w_t}-\nabla_{w}f(w ,z_{J_{t,i}})|_{w=w'_t} \Vert \numberthis \label{eq:second_order_terms_boundmb}\\
&\leq  (1+\beta\alpha_t) \delta_t   +  \frac{2\alpha_t }{mK}\sum^m_{i=1}\sum^K_{k=1} \frac{\mu \beta }{2} \E_{\mathbf{U}_{t,i,k}}\left[ \Vert U^t_{k,i}\Vert ^3 \right] + \alpha_t  \sqrt{\frac{3d-1}{K}}  \beta \delta_t \numberthis \label{eq:expansive_smoothmb}\\
 &\leq \lp 1+ \beta\alpha_t\MDK \rp \delta_t   + \mu \beta\alpha_t  (3+d)^{3/2}, \numberthis \label{eq:iddUmb}
\end{align*} to find the inequality \eqref{eq:Lemma1ongradientdifferencesmb} we applied Lemma \ref{lemma:variance_reduction_zero_mean}, inequality \eqref{eq:second_order_terms_boundmb} comes from the triangle inequality and $\beta$-smoothness, to derive the inequality \eqref{eq:expansive_smoothmb} we applied the $1+\beta\alpha_t$-expansive property for the $G_{\!J_t} (\cdot)$ mapping (Lemma \ref{lemma:Mini-Batch SGD Growth Recursion}) and the $\beta$-smoothness of the loss function, finally the inequality \eqref{eq:iddUmb} holds since the random vectors $U^t_{k,i}\sim\mc{N}(0,I_d)$ are i.i.d. and $\E[\Vert U^t_{k,i}\Vert]\leq (3+d)^{3/2}$ for all $k\in \{1,2,\ldots,K\}$, $i\in \{1,2,\ldots,m\}$ and $t\leq T$. Under the choice $\mu\leq cL\MDK/(n\beta (3+d)^{3/2}) $, \eqref{eq:iddUmb} gives the first part the inequality in Lemma \ref{lemma:ZoSS Growth Recursion mb}.

\noindent We continue by considering 
%
%
the event $\tilde{\mathcal{E}}^c_t\triangleq \{ G_{\!J_t}(\cdot) \neq G'_{\!J'_t}(\cdot)\}$. Recall that $\tilde{\mc{E}}_t^c$ occurs with probability $\P(\tilde{\mathcal{E}}_t)=m/n$ for all $t\leq T$. Under the event $\tilde{\mc{E}}_t^c$ similarly to \eqref{eq:W_difference_Ec} we derive the difference\begin{align*}
    &\tilde{G}_{\!J_t}(w_t) - \tilde{G}'_{\!J'_t}(w'_{t}) 
    \\& = \underbrace{w_t-\frac{\alpha_t}{m}\sum^m_{i=1}  \nabla_{w}f(w ,z_{J_{t,i}})|_{w=w_t}}_{G_{\!J_t}(w_t)} - \bigg( \underbrace{w'_t  -\frac{\alpha_t}{m}\sum^m_{i=1}\nabla_{w}f(w ,z'_{J'_{t,i}})|_{w=w'_t}}_{G_{\!J_t}'(w'_t)} \bigg) \\&\quad - \frac{\alpha_t}{m K}\sum^m_{i=1}\sum^K_{k=1} \lp \frac{\mu}{2} U^{\text{T}}_k \nabla^2_{w}f(w ,z_{J_{t,i}})|_{w=W^*_{k,t,i}} U^t_{k,i} \rp U^t_{k,i}  \\&\quad  + \frac{\alpha_t}{mK}\sum^m_{i=1}\sum^K_{k=1} \lp \frac{\mu}{2} U^{\text{T}}_k \nabla^2_{w}f(w ,z'_{J'_{t,i}})|_{w=W^\dagger_{k,t,i}} U^t_{k,i}\rp U^t_{k,i}  \\ &    \\&\quad-   \frac{\alpha_t}{m}\sum^m_{i=1}\bigg( \frac{1}{K}\sum^K_{k=1} \inp{\nabla_{w}f(w ,z_{J_{t,i}})|_{w=w_t}-\nabla_{w}f(w ,z'_{J'_{t,i}})|_{w=w'_t}}{ U^t_{k,i}}U^t_{k,i} \\&\quad\quad\quad- ( \nabla_{w}f(w ,z_{J_{t,i}})|_{w=w_t}-\nabla_{w}f(w ,z'_{J'_{t,i}})|_{w=w'_t})\bigg).\numberthis\label{eq:min_batch__diff_2}
\end{align*} By using the triangle inequality and Lemma \ref{lemma:variance_reduction_zero_mean} we get\begin{align*}
    & \E [\Vert \tilde{G}_{\!J_t}(w_t) - \tilde{G}'_{\!J'_t}(w'_{t}) \Vert ] \\
    & \leq \Vert G_{\!J_t}(w_t) - G'_{\!J'_t}(w'_t) \Vert  \\& \quad +\E_{\mathbf{U}^{\otimes K \times m }_t}\left[ \left\Vert \frac{\alpha_t}{m K}\sum^m_{i=1}\sum^K_{k=1} \lp \frac{\mu}{2} U^{\text{T}}_k \nabla^2_{w}f(w ,z_{J_{t,i}})|_{w=W^*_{k,t,i}} U^t_{k,i} \rp U^t_{k,i} \right\Vert \right]  \\&\quad+ \E_{\mathbf{U}^{\otimes K \times m }_t} \left[ \left\Vert\frac{\alpha_t}{mK}\sum^m_{i=1}\sum^K_{k=1} \lp \frac{\mu}{2} U^{\text{T}}_k \nabla^2_{w}f(w ,z'_{J'_{t,i}})|_{w=W^\dagger_{k,t,i}} U^t_{k,i}\rp U^t_{k,i}\right\Vert\right]\\ & \quad  +\E_{\mathbf{U}^{\otimes K \times m }_t}\Bigg[ \Bigg\Vert \frac{\alpha_t}{m}\sum^m_{i=1}\bigg( \frac{1}{K}\sum^K_{k=1} \inp{\nabla_{w}f(w ,z_{J_{t,i}})|_{w=w_t}-\nabla_{w}f(w ,z'_{J'_{t,i}})|_{w=w'_t}}{ U^t_{k,i}}U^t_{k,i} \\&\quad\quad\quad- ( \nabla_{w}f(w ,z_{J_{t,i}})|_{w=w_t}-\nabla_{w}f(w ,z'_{J'_{t,i}})|_{w=w'_t})\bigg) \Bigg\Vert\Bigg]\\
&\leq  \Vert G_{\!J_t}(w_t) - G'_{\!J'_t}(w'_t) \Vert + \frac{2\alpha_t }{mK}\sum^m_{i=1}\sum^K_{k=1} \frac{\mu \beta }{2} \E_{\mathbf{U}_{t,i,k}}\left[ \Vert U^t_{k,i}\Vert ^3 \right] \\ & \quad + \frac{\alpha_t}{m}\sum^m_{i=1}\E_{\mathbf{U}^{\otimes K }_{t,i}}\Bigg\Vert \bigg( \frac{1}{K}\sum^K_{k=1} \inp{\nabla_{w}f(w ,z_{J_{t,i}})|_{w=w_t}-\nabla_{w}f(w ,z'_{J'_{t,i}})|_{w=w'_t}}{ U^t_{k,i}}U^t_{k,i} \\&\quad\quad\quad- ( \nabla_{w}f(w ,z_{J_{t,i}})|_{w=w_t}-\nabla_{w}f(w ,z'_{J'_{t,i}})|_{w=w'_t})\bigg) \Bigg\Vert\\
&\leq  \Vert G_{\!J_t}(w_t) - G'_{\!J'_t}(w'_t) \Vert   + \frac{2\alpha_t }{mK}\sum^m_{i=1}\sum^K_{k=1} \frac{\mu \beta }{2} \E_{\mathbf{U}_{t,i,k}}\left[ \Vert U^t_{k,i}\Vert ^3 \right]  \\&\quad + \frac{\alpha_t}{m}\sum^m_{i=1}  \sqrt{\frac{3d-1}{K}} \Vert \nabla_{w}f(w ,z_{J_{t,i}})|_{w=w_t}-\nabla_{w}f(w ,z'_{J'_{t,i}})|_{w=w'_t} \Vert \numberthis \label{eq:second_order_terms_boundmb2}\\
&=  \Vert G_{\!J_t}(w_t) - G'_{\!J'_t}(w'_t) \Vert   + \frac{2\alpha_t }{mK}\sum^m_{i=1}\sum^K_{k=1} \frac{\mu \beta }{2} \E_{\mathbf{U}_{t,i,k}}\left[ \Vert U^t_{k,i}\Vert ^3 \right]  \\&\quad +\sqrt{\frac{3d-1}{K}} \frac{\alpha_t }{m}   \sum^m_{i=1 , i\neq i^*}   \Vert \nabla_{w}f(w ,z_{J_{t,i}})|_{w=w_t}-\nabla_{w}f(w ,z'_{J'_{t,i}})|_{w=w'_t} \Vert  \\&\quad +\sqrt{\frac{3d-1}{K}} \frac{\alpha_t }{m}      \Vert \nabla_{w}f(w ,z_{J_{t,i^*}})|_{w=w_t}-\nabla_{w}f(w ,z'_{J'_{t,i^*}})|_{w=w'_t} \Vert \\
&=  \Vert G_{\!J_t}(w_t) - G'_{\!J'_t}(w'_t) \Vert   + \frac{2\alpha_t }{mK}\sum^m_{i=1}\sum^K_{k=1} \frac{\mu \beta }{2} \E_{\mathbf{U}_{t,i,k}}\left[ \Vert U^t_{k,i}\Vert ^3 \right]  \\&\quad +\sqrt{\frac{3d-1}{K}} \frac{\alpha_t }{m}   \sum^m_{i=1 , i\neq i^*}   \Vert \nabla_{w}f(w ,z_{J_{t,i}})|_{w=w_t}-\nabla_{w}f(w ,z_{J_{t,i}})|_{w=w'_t} \Vert  \\&\quad +\sqrt{\frac{3d-1}{K}} \frac{\alpha_t }{m}      \Vert \nabla_{w}f(w ,z_{J_{t,i^*}})|_{w=w_t}-\nabla_{w}f(w ,z'_{J'_{t,i^*}})|_{w=w'_t} \Vert  \numberthis \label{eq:second_order_terms_boundmb4}\\
&\leq  \Vert G_{\!J_t}(w_t) - G'_{\!J'_t}(w'_t) \Vert   + \frac{2\alpha_t }{mK}\sum^m_{i=1}\sum^K_{k=1} \frac{\mu \beta }{2} \E_{\mathbf{U}_{t,i,k}}\left[ \Vert U^t_{k,i}\Vert ^3 \right]  \\&\quad +\sqrt{\frac{3d-1}{K}} \frac{\alpha_t }{m}   (m-1)\beta \delta_t  +\sqrt{\frac{3d-1}{K}} \frac{\alpha_t }{m} 2L  \numberthis \label{eq:second_order_terms_boundmb5}\\
&\leq \lp 1+\frac{m-1}{m} \beta \alpha_t \rp  \delta_t + \frac{2}{m} L \alpha_t   +  \mu \beta\alpha_t (3+d)^{3/2}  \\&\quad +\sqrt{\frac{3d-1}{K}} \frac{\alpha_t }{m}   (m-1)\beta \delta_t  +\sqrt{\frac{3d-1}{K}} \frac{\alpha_t }{m} 2L  \numberthis \label{eq:second_order_terms_boundmb6}\\
&= \lp 1+\frac{m-1}{m} \beta \alpha_t\MDK \rp  \delta_t  + \frac{2L \alpha_t}{m} \MDK  +  \mu \beta\alpha_t (3+d)^{3/2},   \numberthis \label{eq:second_order_terms_boundmb7}
\end{align*} we find the inequality \eqref{eq:second_order_terms_boundmb2} by applying Lemma \ref{lemma:variance_reduction_zero_mean}, the inequality \eqref{eq:second_order_terms_boundmb4} holds since $z_{J_{t,i}} = z'_{J'_{t,i}}$ for any $i\neq i^*$, we find \eqref{eq:second_order_terms_boundmb5} by using the triangle inequality and $\beta$-smoothness (for $i\neq i^*$) and $L-$Lipschitz condition to bound the norm of the gradients $\nabla_{w}f(w ,z_{J_{t,i^*}})|_{w=w_t}$ and $\nabla_{w}f(w ,z'_{J'_{t,i^*}})|_{w=w'_t}$. In \eqref{eq:second_order_terms_boundmb6} we apply Lemma \ref{lemma:Mini-Batch SGD Growth Recursion} to bound the quantity $ \Vert G_{\!J_t}(w_t) - G_{\!J_t}'(w'_t) \Vert$. Under the selection of $\mu\leq cL\MDK/(n\beta (3+d)^{3/2}) $, Eq. \eqref{eq:second_order_terms_boundmb7} gives the second part of the inequality in Lemma \ref{lemma:ZoSS Growth Recursion mb}. \qedwhite

\paragraph{Proof of Theorem \ref{thm:recur_solution_nonconvex_mb}} We apply Lemma \ref{lemma:ZoSS Growth Recursion mb} to get
\begin{align*}
    &\E[\delta_{t+1}|\mc{E}_{\delta_{t_0}}] \\&= \P (\mc{E}_t)\E[\delta_{t+1}|\mc{E}_t,\mc{E}_{\delta_{t_0}}] + \P (\mc{E}^c_t)\E[\delta_{t+1}|\mc{E}^c_t,\mc{E}_{\delta_{t_0}}] \\
    &=\lp 1-\frac{m}{n}\rp\E[\delta_{t+1}|\mc{E}_t,\mc{E}_{\delta_{t_0}}] + \frac{m}{n}\E[\delta_{t+1}|\mc{E}^c_t ,\mc{E}_{\delta_{t_0}}]\\
    &= \lp 1-\frac{m}{n}\rp \lp \lp 1+ \beta\alpha_t\MDK \rp \E[\delta_{t}|\mc{E}_{\delta_{t_0}}]   + \frac{cL\alpha_t}{n}\MDK \rp\\
    &\quad + \frac{m}{n}   \lp 1+\frac{m-1}{m} \beta \alpha_t\MDK \rp  \E[\delta_{t}|\mc{E}_{\delta_{t_0}}]   + \frac{m}{n}\frac{2L \alpha_t}{m} \MDK  + \frac{m}{n} \frac{cL\alpha_t}{n}\MDK \\
    &=       \lp 1+ \beta \alpha_t\MDK \lp 1-\frac{1}{n} \rp  \rp                   \E[\delta_{t}|\mc{E}_{\delta_{t_0}}]  +        \frac{2L \alpha_t}{n} \MDK           +   \frac{cL\alpha_t}{n}\MDK.\numberthis \label{eq:growth_minibatch}
\end{align*} The last display characterizes the general case of nonconvex loss and coincides with the inequality \eqref{eq:recusrion} (since $\eta =1+\beta \alpha_t$). As a consequence the solution of the recursion \eqref{eq:growth_minibatch} is \begin{align}
    \E[\delta_T|\mc{E}_{\delta_{t_0}}] \leq \frac{(2+c)L\MDK}{n}  \sum^T_{t=t_0+1} \alpha_t \prod^T_{j=t+1}\lp 1+ \beta \alpha_j \MDK \lp 1-\frac{1}{n} \rp  \rp.\label{eq:Recur_sol_R_mb}
\end{align} The last display completes the proof. \qedwhite 
\section{Full-Batch GD}\label{Append_GD}
As a byproduct of our analysis we derive generalization error bounds for the full-batch gradient decent. Although our results reduce to the full-batch GD by a direct calculation of the limits $c\rightarrow 0$, $K\rightarrow \infty$ and setting the batch size $m$ equal to $n$, we separately prove generalization error bounds for the full-batch GD for clarity. 

\begin{corollary}\label{corollary_GD}
[\textbf{Stability and Generalization Error of Full-Batch GD | Nonconvex Loss}]\label{thm:recur_solution_nonconvex_GD}
Assume that the loss function $f(\cdot, z)$ is $L$-Lipschitz and $\beta$-smooth for all $z\in \mc{Z}$. Consider the (deterministic) full-batch GD algorithm, initial state $W_0 =W'_0$, iterates $W_t = G_{\! S}(W_t) $, $W'_t = G'_{\! S'}(W'_t) $ for $t>0$, and with final-iterate estimates $W_T$ and $W'_T$ corresponding to the data-sets $S,S'$, respectively (that differ in exactly one entry). Then the discrepancy $\delta_T \triangleq \Vert  W_{T} - W'_{T} \Vert$ satisfies the inequality
\begin{align}
   \!\! \delta_{T} \leq \frac{2L}{n}\sum^T_{t=1} \alpha_t \prod^T_{j=t+1} \lp 1+\frac{n-1}{n} \alpha_j \beta \rp.\label{eq:delta_bounds_GD}
\end{align} Further if $\alpha_t\leq C/t$ for any $t>0$ and some $C>0$ then \begin{align}
   \!\!\!\! |\eps_{\text{gen}}|=|\E_S [\E_z [f(W_T , z)] - \frac{1}{n}\sum_{z\in S}  f(W_T,z) ]|\leq \frac{2L^2\lp e T \rp^{C\beta}}{n}\min\left\{C+\beta^{-1},C\log (eT)\right\}.
\end{align}
\end{corollary} The proof of Corollary \ref{thm:recur_solution_nonconvex_GD} follows.
\paragraph{Proof of Corollary \ref{thm:recur_solution_nonconvex_GD} (Full-Batch GD)}\label{Appendix_proof_Full-batch GD}
In the case of full-batch GD the algorithm is deterministic and we assume that $z_1, z_2,\ldots, z_i,\ldots,z_n , z'_i$ are i.i.d. and define $S\triangleq(z_1, z_2,\ldots, z_i,\ldots,z_n)$ and $S'\triangleq(z_1, z_2,\ldots, z'_i,\ldots,z_n)$, $W_0 = W'_0$, the updates for any $t\geq 1$ are \begin{align}
  W_{t+1} &=  W_t- \frac{\alpha_t}{n} \sum^n_{j=1} \nabla f(W_t ,z_{j}),\\
  W'_{t+1} &=  W'_t- \frac{\alpha_t}{n} \sum^n_{j=1, j\neq i } \nabla f(W'_t ,z_{j}) - \frac{\alpha_t}{n} \nabla f(W'_t ,z'_{i}).
\end{align} Then for any $t\geq 1$ \begin{align*}
    &\delta_{t+1}\\& \leq\delta_t +\frac{\alpha_t}{n} \bigg\Vert \sum^n_{j=1, j\neq i } \nabla f(W_t ,z_{j})- \nabla f(W'_t ,z_{j}) \bigg\Vert + \frac{\alpha_t}{n}\Vert \nabla f(W_t ,z_{i})- \nabla f(W'_t ,z'_{i}) \Vert\\
    &\leq \delta_{t} + \frac{\alpha_t (n-1)}{n} \beta \delta_{t} +  \frac{2L \alpha_t}{n}\\
    & = \lp 1 + \frac{ (n-1)}{n} \beta \alpha_t \rp \delta_{t} +  \frac{2L \alpha_t}{n}.
\end{align*} Then by solving the recursion we find \begin{align}
    \delta_{T} \leq \frac{2L}{n}\sum^T_{t=1} \alpha_t \prod^T_{j=t+1} \lp 1+\frac{n-1}{n} \alpha_j \beta \rp.
\end{align}Under the choice $\alpha_t \leq C/t$ the last display gives \begin{align*}
    \delta_{T} &\leq \frac{2L}{n}\sum^T_{t=1} \frac{C}{t}\prod^T_{j=t+1} \lp 1+\frac{n-1}{n} \frac{C}{j} \beta \rp\\
    &\leq \frac{2L}{n}\sum^T_{t=1} \frac{C}{t}\prod^T_{j=t+1} \lp 1+ \frac{C}{j} \beta \rp\\
    &\leq \frac{2L}{n}\sum^T_{t=1} \frac{C}{t}\prod^T_{j=t+1} \exp \lp \frac{C}{j} \beta \rp \\
    & = \frac{2L}{n}\sum^T_{t=1} \frac{C}{t} \exp\lp \sum^T_{j=t+1}\frac{C}{j} \beta \rp\\
    &\leq \frac{2L}{n}\sum^T_{t=1} \frac{C}{t} \exp\lp C\beta \lp 1+ \log\frac{T}{t+1}  \rp \rp\\
    & = \frac{2L\lp e T \rp^{C\beta}}{n}\sum^T_{t=1} \frac{C}{t}\frac{1}{(t+1)^{C\beta}} \\
    &\leq \frac{2L\lp e T \rp^{C\beta}}{n}\sum^T_{t=1} \frac{C}{t^{C\beta+1}} \\
    &\leq \frac{2CL\lp e T \rp^{C\beta}}{n}\min\left\{\frac{C\beta+1}{C\beta},\log (eT)\right\}.\label{eq:DG_Ws_sensitivity}\numberthis
\end{align*} Then \begin{align}
   |\E_{S} [ R_S (A_S)  -   R(A_S) ]| &=  | \E_{S,z'_i}[ f(W_T,z'_i) - f(W'_T,z'_i) ]|\label{eq:BELemma7} \\&\leq \E_{S,z'_i}[\vert f(W_T,z'_i) - f(W'_T,z'_i)\vert ]\nonumber\\
   &   \leq L  \E_{S,z'_i} \Vert W_{T} - W'_{T} \Vert \label{eq:Lipschitz_GDproof}\\
   &\leq \frac{2L^2\lp e T \rp^{C\beta}}{n}\min\left\{C+\beta^{-1},C\log (eT)\right\}.\label{eq:GD_gen_bound_proof}
\end{align} In the above, Eq. \eqref{eq:BELemma7} follows from ~\cite[Lemma 7]{bousquet2002stability},  the inequality \eqref{eq:Lipschitz_GDproof} holds under the Lipschitz property of the loss $f(\cdot,z)$ for any $z$. Finally, we find the last inequality \eqref{eq:GD_gen_bound_proof} by applying the bound in \eqref{eq:DG_Ws_sensitivity}. $\qedwhite$

\section{Excess Risk --- Convex Loss}
Define the time average estimated parameters 
\begin{align}
   \bar{W}_T = \frac{1}{\sum^T_{t=1} \alpha_t} \sum^T_{t=1} \alpha_t W_t,
\end{align}then 
\begin{align}
 \E [\Vert \bar{W}_T -\bar{W}'_T \Vert | \mc{E}_{\delta_{t_0}} ]& \leq \frac{1}{\sum^T_{t=1} \alpha_t} \sum^T_{t=1} \alpha_t \E [ \Vert W_t - W'_t \Vert |\mc{E}_{\delta_{t_0}} ] \\&= \frac{1}{\sum^T_{t=1} \alpha_t} \sum^T_{t=1} \alpha_t\E [ \delta_t |\mc{E}_{\delta_{t_0}} ] \\&\leq \frac{1}{\sum^T_{t=1} \alpha_t} \sum^T_{t=1} \alpha_t \E [ \delta_T |\mc{E}_{\delta_{t_0}} ] = \E [ \delta_T |\mc{E}_{\delta_{t_0}} ].\label{average_stab1}
\end{align} The L-Lipschitz property of the loss and the inequality \eqref{average_stab1} give \begin{align}
    \overline{\eps_{\mathrm{gen}}} \triangleq | \E[ f(\bar{W}_T,z'_i) - f(\bar{W}'_T,z'_i) ]|\leq   \E[| f(\bar{W}_T,z'_i) - f(\bar{W}'_T,z'_i) |]\leq L \E [ \delta_T |\mc{E}_{\delta_{t_0}} ].\label{eq:bargen}
\end{align}  Additionally, since the loss function is convex it is true that,
\begin{align}
    \overline{\eps_{\mathrm{opt}}} \triangleq \E [ R( \bar{W}_T) ] -R(w^*) &\leq \frac{1}{\sum^T_{t=1} \alpha_t} \sum^T_{t=1} \alpha_t \lp \E [ R( W_t ) ] -R(w^*)   \rp\nonumber \\
    &\leq  \frac{1}{\sum^T_{t=1} \alpha_t} \lp \frac{1}{2} \Vert W_0 -W^* \Vert^2 +\frac{d+4}{2} L \sum^T_{t=1} \alpha^2_t \rp.\label{eq:ebaropt}
\end{align} If $\Vert W_0 -W^* \Vert^2 \leq R$, $K=1$, then we may choose \begin{align}
    \alpha_t =\frac{C R}{L\sqrt{3d-1} t}.\label{eq:lr}
\end{align} From \eqref{eq:ebaropt} and \eqref{eq:lr} we find \begin{align}
    \overline{\eps_{\mathrm{opt}}}&\leq \frac{L \sqrt{3d-1}}{C R\log(T+1)} \lp \frac{ R^2}{2} + \frac{d+4}{2} L \lp \frac{C^2 R^2}{L^2 (3d-1) } \rp \frac{\pi^2}{6}  \rp\nonumber\\
    &\leq \frac{RL \sqrt{3d-1}}{2C \log(T+1)} \lp 1 +    \frac{C^2 }{L  }    \rp.
\end{align} Further, inequality \eqref{eq:bargen}, the choice of learning rate in \eqref{eq:lr} and Lemma \ref{thm:recur_solution_convex} give \begin{align}
     \overline{\eps_{\mathrm{gen}}} &\leq \frac{1+\sqrt{3d-1}}{\sqrt{3d-1}} \frac{   (eT)^{CR \beta /L} (2+c)L^2}{n   } \min\left\{ \frac{CR\beta/L+1}{\beta}, \frac{CR}{L}\log (e T) \right\}\\
     &\leq  \frac{ 2  (eT)^{CR \beta /L} (2+c)L^2}{n   } \min\left\{ \frac{CR\beta/L+1}{\beta}, \frac{CR}{L}\log (e T) \right\}.
\end{align} If $C\leq L/2R\beta$, then \begin{align}
     \overline{\eps_{\mathrm{gen}}} &\leq  \frac{ 2  \sqrt{eT} (2+c)L^2}{n   } \min\left\{ \frac{3}{2\beta}, \frac{1}{2\beta}\log (e T) \right\}\\
     &\leq \frac{ 3  \sqrt{eT} (2+c)L^2/\beta}{n   }.
\end{align} We conclude that \begin{align}
    \overline{\eps_{\mathrm{excess}}}\leq \overline{\eps_{\mathrm{gen}}} + \overline{\eps_{\mathrm{opt}}} \leq \frac{ 3  \sqrt{eT} (2+c)L^2/\beta}{n   } + \frac{\sqrt{3d-1} R^2 \beta }{ \log(T+1)} \lp 1 +    \frac{L }{4 R^2 \beta^2  }    \rp.
\end{align}

\end{document}